%% file: colm2026_conference.tex
\documentclass{article} 
\usepackage[preprint]{colm2026_conference}

\usepackage{microtype}
\usepackage{hyperref}
\usepackage{url}
\usepackage{booktabs}

\usepackage{lineno}

\input{math_commands}

\usepackage{amsmath}
\usepackage{amssymb}
\usepackage{mathtools}
\usepackage{amsthm}
\usepackage{adjustbox}
\usepackage{float}
\usepackage{xcolor}
\input{macros}
\input{packages}

\definecolor{darkblue}{rgb}{0, 0, 0.5}
\hypersetup{colorlinks=true, citecolor=darkblue, linkcolor=darkblue, urlcolor=darkblue}

\title{\ours: Transformer with Temporal Middle-Layer Recurrence}

\author{Ziyang Cai\thanks{Joint first authors; equal contribution.}, \,
Xingyu Zhu$^*$, \,
Yihe Dong\thanks{Core contribution.}, \,
Yinghui He, \,
Sanjeev Arora \\
Princeton Language and Intelligence, Department of Computer Science\\
Princeton University \\
\texttt{\{zc5794, xingyu.zhu\}@princeton.edu}
}

\begin{document}

\ifcolmsubmission
\linenumbers
\fi

\maketitle
\begin{abstract}
\input{Sections/abstract}
\end{abstract}

\input{Sections/intro}

\input{Sections/arch.tex}

\input{Sections/expressivity_seperation}

\input{Sections/experiments}



\input{Sections/mechanistic_interpretation}

\input{Sections/related_works}

\input{Sections/discussion}

\section*{Contributions}

Ziyang Cai$^*$ designed the fusion gate, completed the majority of the experiments including data generation and training, and contributed to writing. Xingyu Zhu$^*$ proposed the idea of \ours, wrote the core training and inference script, and contributed to writing. Yihe Dong$^\dagger$ contributed to general discussion and experiments on the architecture design, completed the future token prediction task and part of pre-training experiments, and contributed to writing. Yinghui He worked on the HotpotQA experiment. Sanjeev Arora advised the project. ($^*$Equal contribution; $^\dagger$core contribution.)

\section*{Acknowledgments}
We acknowledge the support from NSF, Schmidt Foundation, DARPA AIQ Program, OpenAI and Google Inc. Ziyang Cai and Xingyu Zhu are additionally supported by the Gordon Y.S. Wu Fellowship in Engineering.

We thank Abhishek Panigrahi for the initial discussion with Xingyu Zhu which motivated this idea. We would also like to thank Yun Cheng, Haoyu Zhao, Liam Fowl, Narutatsu Ri, and Zixuan Wang for helpful discussions during various stages of the project.
\newpage
\bibliography{colm2026_conference}
\bibliographystyle{colm2026_conference}

\appendix
\input{Appendix/Sections/appendix}

\end{document}

%% file: math_commands.tex
\usepackage{amsmath,amsfonts,bm}

\def\eqref#1{equation~\ref{#1}}

\def\1{\bm{1}}

\def\vh{{\bm{h}}}

\def\mH{{\bm{H}}}

\def\mW{{\bm{W}}}

\DeclareMathAlphabet{\mathsfit}{\encodingdefault}{\sfdefault}{m}{sl}
\SetMathAlphabet{\mathsfit}{bold}{\encodingdefault}{\sfdefault}{bx}{n}

\newcommand{\R}{\mathbb{R}}

\newcommand{\sigmoid}{\sigma}

\newcommand{\rbr}[1]{\left(#1\right)}
\newcommand{\srbr}[1]{(#1)}
\newcommand{\sbr}[1]{\left[#1\right]}
\newcommand{\cbr}[1]{\left\{#1\right\}}

\newcommand{\agbr}[1]{\left\langle #1 \right\rangle}

%% file: macros.tex
\usepackage{xcolor}
\usepackage[most]{tcolorbox}

\newcommand{\ours}{T$^2$MLR\xspace} 
\newcommand{\Fuse}{\Phi}
\newcommand{\Fl}[1]{\mathcal{F}_{#1}}

\newcommand{\lstart}{\ell_{\text{start}}}
\newcommand{\lend}{\ell_{\text{end}}}
\newcommand{\dkv}{d_{\text{KV}}}
\newcommand{\rcache}{\mathbf{R}}
\newcommand{\cache}[2]{\mathbf{KV}_{#1}^{(#2)}}

\newcommand{\Wrec}{\mW_{\text{rec}}}
\newcommand{\fcur}{f_{\text{cur}}}
\newcommand{\frec}{f_{\text{rec}}}
\newcommand{\gcur}{\gamma_{\text{cur}}}
\newcommand{\grec}{\gamma_{\text{rec}}}

\newcommand{\alphabet}{\aleph}

\newcommand{\dforward}{d_{\text{forward}}}
\newcommand{\dbackward}{d_{\text{backward}}}

%% file: packages.tex
\usepackage[utf8]{inputenc} 
\usepackage[T1]{fontenc}    
\usepackage[ruled,vlined, linesnumbered]{algorithm2e}
\DontPrintSemicolon

\usepackage{url}            
\usepackage{booktabs}       
\usepackage{amsfonts}       
\usepackage{nicefrac}       
\usepackage{microtype}      
\usepackage{xcolor}         

\usepackage[american]{babel}
\usepackage[utf8]{inputenc} 
\usepackage[T1]{fontenc}    
\usepackage{csquotes}
\usepackage{hyperref}       
\usepackage{url}            
\usepackage{booktabs}       
\usepackage{amsfonts}       
\usepackage{nicefrac}       
\usepackage{microtype}      
\usepackage{xcolor}         
\usepackage{array}          
\usepackage{tikz}
\usepackage{lipsum}

\usetikzlibrary{calc,positioning,decorations.pathreplacing,shapes}
\usepackage{amsmath}
\usepackage{amsthm}
\usepackage{amssymb}
\PassOptionsToPackage{normalem}{ulem}
\usepackage{ulem}
\usepackage{dsfont}
\usepackage[noabbrev, capitalise]{cleveref}

\usepackage{enumitem}
\usepackage{xspace}
\usepackage{graphicx}
\usepackage{subcaption}

\usepackage{wrapfig} 

\theoremstyle{plain}

  \theoremstyle{plain}
  
  \theoremstyle{plain}
  
  \theoremstyle{remark}
  
  \theoremstyle{plain}
  
  \theoremstyle{plain}
  
  \theoremstyle{plain}
  
  \theoremstyle{plain}

\usepackage{mathtools}

\usepackage{contour}
\usepackage[normalem]{ulem}

\crefname{assumption}{Assumption}{Assumption}

\crefname{example}{Example}{Example}

\crefname{definition}{Definition}{Definition}

\makeatother
\crefformat{equation}{(#2#1#3)}

\numberwithin{equation}{section}

\contourlength{0.8pt}

\renewcommand{\underline}[1]{%
  \uline{\phantom{#1}}%
  \llap{\contour{white}{#1}}%
}

\definecolor{OliveGreen}{rgb}{0,0.6,0}
\definecolor{JaumeBlue}{rgb}{0,0,0.6}

\usepackage{babel}
  \providecommand{\conjecturename}{Conjecture}
  \providecommand{\corollaryname}{Corollary}
  \providecommand{\lemmaname}{Lemma}
  \providecommand{\questionname}{Question}
  \providecommand{\propositionname}{Proposition}
  \providecommand{\remarkname}{Remark}
\providecommand{\theoremname}{Theorem}

\newtheorem*{theorem*}{Theorem}


%% file: Sections/abstract.tex


Transformer reasoning is limited by autoregressive decoding, which repeatedly compresses rich hidden computation through token space and makes it difficult for intermediate reasoning states to persist across time. We introduce Transformers with Temporal Middle-Layer Recurrence (\ours), a transformers-based latent reasoning architecture that fuses a cached middle-layer representation from the previous token directly into an earlier layer of the current token position, enabling abstract intermediate computation to persist across decoding steps with little inference overhead.
Across natural-language pretraining and multi-hop reasoning finetuning, \ours consistently outperforms data- and parameter-matched Transformer baselines. Moreover, applying recurrence to only a localized middle-layer block (as little as 20\% of the network) often outperforms full-layer recurrence. Importantly, \ours does not require pretraining from scratch: retrofitting the recurrent pathway into an existing pretrained 1.7B Transformer and briefly finetuning substantially improves math reasoning, lowering the barrier to practical adoption. These results suggest that effective latent reasoning in Transformers does not require looping over all layers as in previous works, but can instead emerge more strongly from targeted middle-layer recurrence.



%% file: Sections/intro.tex
\section{Introduction}

Recent developments in large language models have demonstrated remarkable capabilities in reasoning, solving complex problems in mathematics, physics, and other scientific domains \citep{wei2022chainofthought, cobbe2021gsm8k, openai2023gpt4}. Many of these tasks require multi-step reasoning in which intermediate abstract reasoning states must be iteratively refined before producing a final answer.
Despite these advances, the underlying Transformer architecture \citep{vaswani2017attention} remains fundamentally token-centric: auto-regressive generation repeatedly projects rich, high-dimensional latent representations back into a sparse one-hot vector in the token space at each decoding step, and this discrete representation is then used as the sole input for the next forward pass. This repeated projection creates an information bottleneck that limits how latent intermediate reasoning states can persist across time and influence the computation for future tokens.

\begin{figure}[t]
    \centering
    \includegraphics[width=0.95\linewidth]{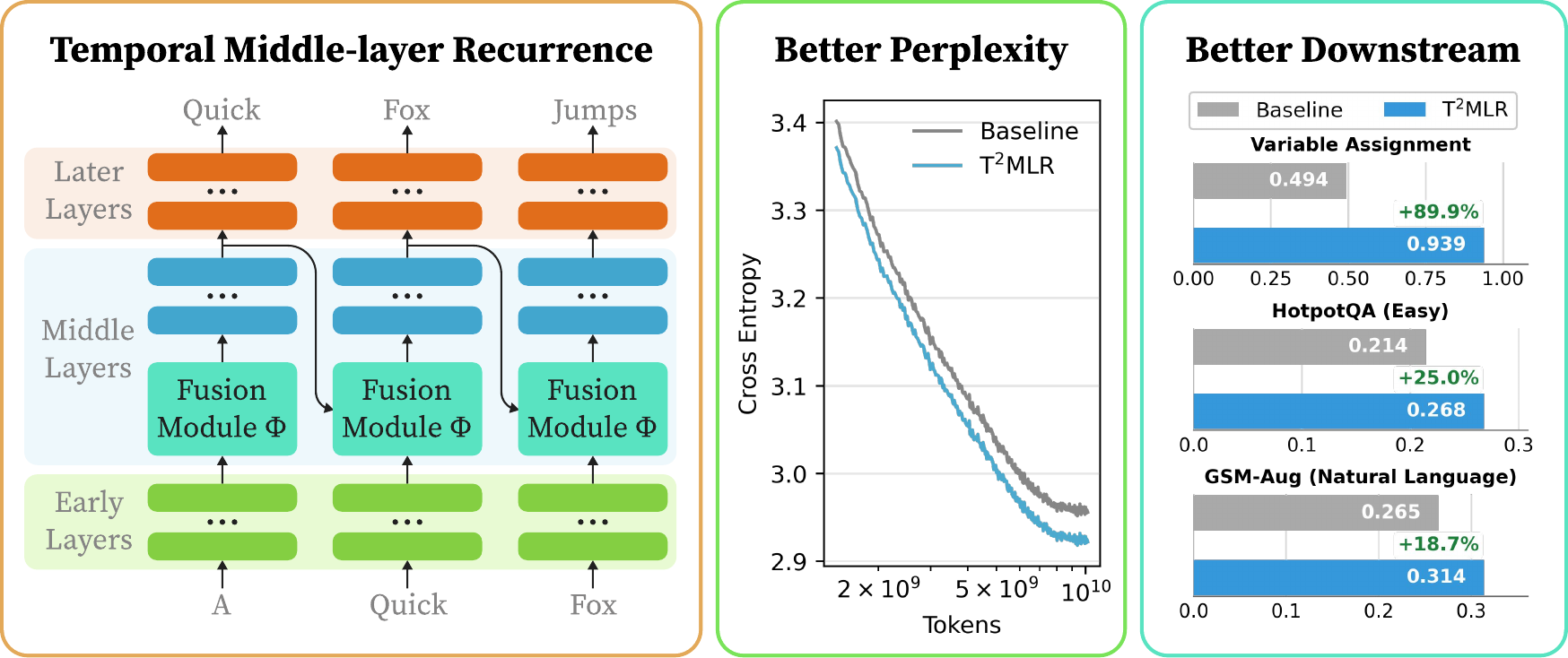}
    \caption{\ours fuses representation from a deep layer from the previous token position into a shallow layer of the current token position (left). It gets better pretraining perplexity (see \cref{sec:exp-pretraining}) and reasoning downstream performance (see \cref{sec:multi-hop-datasets} )}
    \label{fig:teaser}
    \vspace{-0.18in}
\end{figure}

To further improve the reasoning capability of Transformers, a growing line of work on \emph{latent reasoning} seeks to relax this temporal constraint by enabling computation to persist in continuous latent space beyond explicit token-level intermediates. Such approaches either perform reasoning entirely in continuous space \citep{hao2024coconut, shen2025codi}, or propagate uncertainty by forming linear combinations of token embeddings induced by post-softmax distributions \citep{zhang2025soft, zhuang2025mixture, yue2025hybridlatentreasoningreinforcement, tang2026multiplexthinkingreasoningtokenwise}. However, these methods typically operate on representations at or beyond the final layer and feed recurrent signals into the next token through the input embedding. As a result, recurrent information is forced to live outside the middle layers, where abstract reasoning is known to occur more prominently \citep{tenney2019bert, geva2021transformer, meng2022locating, saunshi2024inductive,atanas2025modulardatasetdemonstratellm}.

Concurrently, another line of work has attempted to scale reasoning capacity by extending computation along the depth dimension. These approaches loop over the transformer layers multiple times during the forward pass for a single token without increasing parameter count \citep{saunshi2025latent, geiping2025recurrentdepth,zhu2025ouro}. While successful at amplifying reasoning capabilities, such architectures incur increased inference cost which scales with the number of loops.

In this paper, we introduce Transformers with Temporal Middle-Layer Recurrence (\ours),\footnote{Code: \url{https://github.com/princeton-pli/T2MLR}} a novel Transformer-based latent-reasoning architecture addressing these limitations through middle-layer temporal recurrence. Instead of confining recurrence to the token or embedding space, or increasing inference-time depth via looping, \ours allows abstract intermediate representations computed in the middle layers of the network to persist and evolve across decoding steps.
Concretely, we inject representations from a deeper layer at the previous token directly into an earlier layer of the current token via a gated recurrent pathway. This design enables temporally extended latent reasoning while preserving standard auto-regressive decoding, dense token-level supervision during training, and the inference-time computational profile of a standard transformer.

A central practical challenge for latent-reasoning architectures is scalable teacher-forced training: recurrent latent dependencies across decoding steps often break standard sequence-parallel training, making pretraining and finetuning much less scalable. In this work, we address it with an approximate temporal-parallel training scheme (\cref{sec:bfa}). To the best of our knowledge, this is the first continuous chain-of-thought variant pretrained with dense teacher forcing under scalable sequence parallelism.

Empirically, we show that on data and parameter matched settings, \ours yields substantial gains across tasks that stress different aspects of reasoning. The main findings are summarized as follows:

\begin{itemize}[leftmargin=2em]
    \item Shallow \ours solves \textsc{S5-Retrieval}, a challenging synthetic benchmark requiring both non-solvable group state tracking and in-context retrieval, where standard Transformers and recurrent models fail in isolation (Section~\ref{sec:s5-result}).
    \item In pretraining settings, \ours achieves lower perplexity against parameter-matched Transformers and demonstrates clear improvement on NLP benchmarks (Section~\ref{sec:exp-pretraining}). The best improvement is seen when only looping over 20\% of the middle layers.
    \item Finetuning on downstream reasoning datasets such as GSM-Aug \citep{deng2023implicitcot}, ProsQA \citep{hao2024coconut}, HotPotQA~\citep{yang2018hotpotqa}, and Variable Assignment~\citep{saunshi2024inductive}, \ours consistently outperforms parameter- and data-matched baselines (Sections~\ref{sec:multi-hop-datasets}), and notably, middle layer recurrence variants consistently outperform full-layer recurrence ones.
    \item The gains persist as we scale to 361M and 1B parameters and 50B pretraining tokens (\cref{sec:apdx-scaling}). Crucially, unlike looped and latent-recurrent baselines, \ours adds at most $\sim$8\% per-token inference overhead (\cref{sec:apdx-inference-overhead}), trading additional \emph{training} compute for a low-overhead recurrent latent pathway at inference.
    \item \ours does not require pretraining from scratch: \emph{retrofitting} the recurrent pathway into a pretrained SmolLM2-1.7B-Instruct model and briefly finetuning on math data improves GSM8K accuracy from $35.8$ to $39.9$ and MATH500 from $12.8$ to $18.0$ over the identically-finetuned baseline (\cref{sec:exp-retrofit}), showing the architecture can be adopted at scale without a full pretraining run.
\end{itemize}

%% file: Sections/arch.tex
\section{Transformer with Temporal Middle-Layer Recurrence}
\label{sec:arch}

In this section, we formally introduce our proposed architecture: \ours. We will go over the design motivation, characterize the recurrence and representation fusion module, and describe how we are training the model.

\subsection{Motivation: Effect of Middle Layers and the Information Bottleneck}
\label{sec:motivation}

Recent mechanistic analyses reveal that intermediate layers serve as the primary locus of abstract reasoning in Transformers, whereas early layers focus on lexical and syntactic processing and late layers specialize in projecting representations onto the output vocabulary \citep{tenney2019bert, geva2021transformer, meng2022locating, saunshi2024inductive, atanas2025modulardatasetdemonstratellm}. This suggests that the most valuable reasoning-related computation occurs mainly in the middle layers, and that representations produced after these layers encode rich information about the model’s current reasoning state.

However, autoregressive inference in a standard decoder-only Transformer provides no mechanism for such intermediate representations computed for the previous token to be directly leveraged during the forward pass of the current token: shallow layers processing the current token cannot directly access deeper-layer representations computed at the previous step, even though these representations are already available in memory. To exploit this information, the model must either reconstruct it by traversing the same depth again and retrieving it indirectly via attention, or rely on the input embedding, a highly compressed representation that must pass through the unembed--decode--embed bottleneck before reaching deeper layers.


\subsection{Temporal Middle-Layer Recurrence}

\begin{figure}[b]
    \centering
    \includegraphics[width=0.96\linewidth]{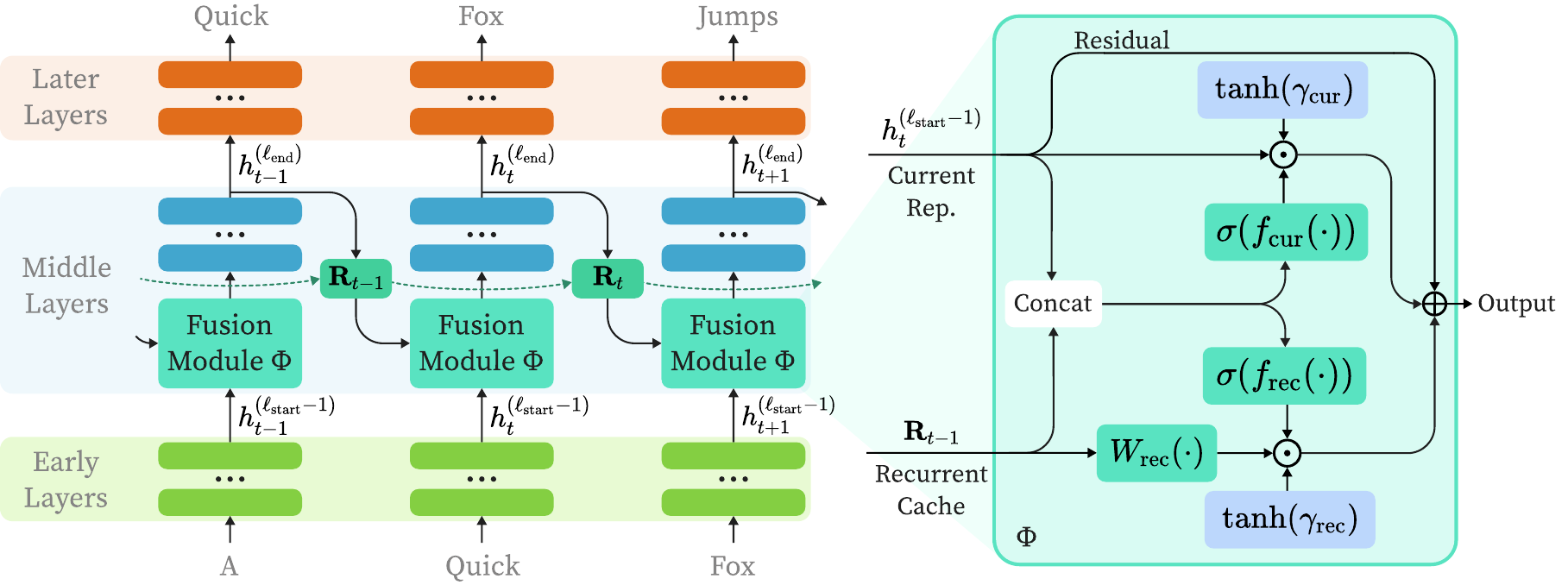}
    \vspace{-0.05in}
    \caption{Illustration of the \ours architecture (left) and the representation fusion module $\Fuse$ (right). When doing forward pass for the $t$-th token, we use the representation after layer $\lend$ to update the recurrent cache $\rcache_{t}$. During the forward pass for the $t+1$-th token, $\rcache_{t}$ is then merged with the representation before layer $\lstart$ via the representation fusion module.}
    \label{fig:main_arch}
    \vspace{-0.1in}
\end{figure}

To relax the bottleneck without changing the autoregressive interface, we introduce a lightweight recurrent pathway that exposes a cached intermediate representation from the previous step as an explicit input at the current step. We now make our architectural change precise in this section.

We start from a standard $L$-layer decoder-only Transformer with hidden dimension $d$, and work at the level of Transformer blocks with KV caches \citep{ott-etal-2019-fairseq}.
We abstract out the key and value caches, and assume each token at each layer corresponds to a single cache vector in $\R^{\dkv}$.
Let $\vh_t^{(0)}\in\R^{d}$ be the input embedding of the $t$-th token $x_t$.
During autoregressive generation, each layer $\ell\in[L]$ maintains a KV cache of past tokens up to step $t-1$, denoted by
$\cache{1:t-1}{\ell} \in \R^{(t-1)\times \dkv}.$
We view the $\ell$-th Transformer block as a map that takes the current token representation together with the past cache and returns an updated representation and an updated cache:
\begin{equation}
\rbr{\vh_t^{(\ell)},\, \cache{1:t}{\ell}}
=\Fl{\ell}\rbr{\vh_t^{(\ell-1)},\, \cache{1:t-1}{\ell}},
\qquad \ell=1,2,\dots,L.
\label{eqn:abstract-transformer-forward}
\end{equation}

To break the depth-time barrier, we introduce \emph{temporal middle-layer recurrence}, parameterized by two layer indices
$1\le \lstart \le \lend \le L$ and a learnable representation fusion module $\Fuse:\R^d\times\R^d\to\R^d$. We additionally maintain a constant-size \emph{recurrent cache} $\rcache_{t}\in\R^d$ for each decoding step $t$. $\rcache_{t}$ will store information for the representation after $\lend$ at the current step and be used by $\lstart$ in the next.
Formally, for any step $t\ge 1$, we modify the $\lstart$-th transformer block computation from \cref{eqn:abstract-transformer-forward} into:
\begin{equation}
\rbr{\vh_t^{(\lstart)},\, \cache{1:t}{\lstart}}
=\Fl{\lstart}\rbr{\Fuse\rbr{\vh_t^{(\lstart-1)},\, \rcache_{t-1}},\, \cache{1:t-1}{\lstart}}.
\label{eqn:tmlr-forward}
\end{equation}
Before layer $\lstart$, we fuse the recurrent cache $\rcache_{t-1}$ with the pre-$\lstart$  representation $\vh_t^{(\lstart-1)}$.
This creates a direct pathway, allowing representations computed via the middle layers in the previous step to be utilized early on in the current step.
The computation for the other layers remains the same as in a standard decoder-only transformer. After passing through layer $\lend$, we will compute the recurrent cache $\rcache_{t}$ based on $\vh_t^{(\lend)}$ and $\rcache_{t-1}$, preparing for the next recurrence.


\subsection{Gated Fusion and Recurrent Cache Update}
\label{sec:gating}

Now we are ready to describe how the recurrent cache is fused with the current stream of information from shallower layers.
Recall that $\vh_t^{(\lstart-1)} \in \R^{d}$ is the representation of the current token
before layer $\lstart$ and $\rcache_{t-1} \in \R^{d}$ is the recurrent cache from
the previous decoding step.
The gated fusion module computes
\begin{equation}
\label{eq:gating}
\begin{aligned}
\Fuse\rbr{\vh_t^{(\lstart-1)},\, \rcache_{t-1}}=\ 
&\vh_t^{(\lstart-1)}\\&+\tanh\rbr{\gcur}\,\sigma\rbr{\fcur\rbr{\sbr{\vh_t^{(\lstart-1)}, \rcache_{t-1}}}}\odot\vh_t^{(\lstart-1)}\\
&+\tanh\rbr{\grec}\,\sigma\rbr{\frec\rbr{\sbr{\vh_t^{(\lstart-1)}, \rcache_{t-1}}}}
\odot\Wrec \rcache_{t-1},
\end{aligned}
\end{equation}
where $\fcur,\frec:\R^{2d}\to\R^d$ are learnable linear layers applied to the concatenation of the
current representation and the recurrent cache,
$\Wrec:\R^d\to\R^d$ is a learnable linear projection, and
$\sigma(\cdot)$ denotes the element-wise sigmoid function.

The scalar factors $\tanh\rbr{\gcur}$ and $\tanh\rbr{\grec}$ act as learnable input-independent gates,
while the element-wise $\sigmoid$ terms provide local, input-dependent modulation. We find this design particularly helpful for early training stability as we could initialize $\gamma$'s to zero and the gates randomly.
The fused representation $\Fuse\srbr{\vh_t^{(\lstart-1)}, \rcache_{t-1}}$ the follows the standard transformers forward pass from $\lstart$ to layer $\lend$ .

After the computation at layer $\lend$, the recurrent cache is updated as
\begin{equation}
\label{eq:cache-update}
\rcache_t
=
\mathrm{RMSNorm}\rbr{\vh_t^{(\lend)}
+
\rcache_{t-1}}.
\end{equation}
The updated cache $\rcache_t$ is used in the next decoding step through the fusion in
\cref{eqn:tmlr-forward}.

\subsection{Approximated Training of \ours with Temporal Parallelism}
\label{sec:bfa}



\begin{figure}[t!]
    \centering
    \includegraphics[width=\linewidth]{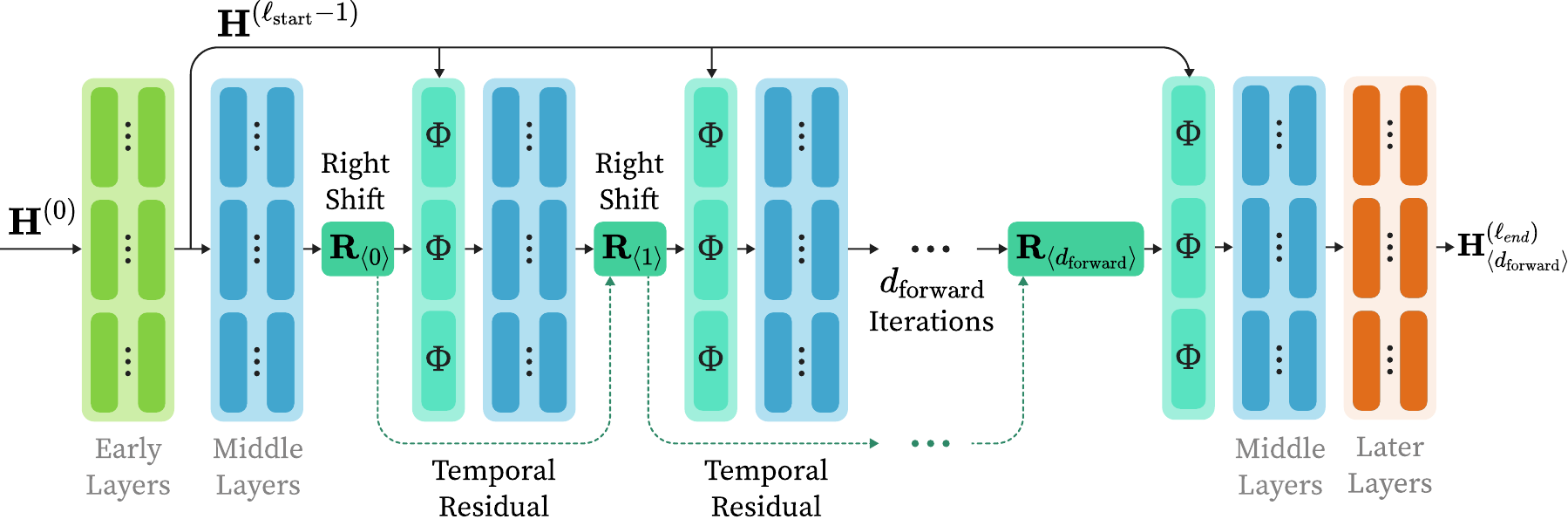}
    \caption{Approximated training scheme of \ours. The ground truth residual cache $\rcache^*$ is approximated by a Jacobi iteration passing through the middle layers for $\dforward$ times.}
    \label{fig:bfa-demo}
\end{figure}

Like all previous transformer-based latent-reasoning architectures which involves recurrence of continuous representations, \ours cannot directly adopt the standard token-parallel training procedure used by vanilla Transformers.
To retain scalability, we approximate $\rcache$ for all tokens in a sequence in parallel using a constant number of Jacobi fixed-point iterations similar to \citep{wu2025parallel} (see illustration in \cref{fig:bfa-demo}).
Due to space constraints, we defer the full training characterization to \cref{alg:bfa_tmlr_minimal} in Appendix \cref{sec:apdx-bfa}, and only provide a simplified description in this section.

We first run a standard forward pass assuming no recurrent cache is available, and take the resulting representation after layer $\lend$ (shifted appropriately) as an initial cache $\rcache^{\agbr{0}}$.
We then fuse this cache with the representation before layer $\lstart$, run the network forward again through the middle layers up to layer $\lend$, and use the new layer-$\lend$ representations to get a refined cache $\rcache^{\agbr{1}}$.
During training, we repeat this procedure for a fixed number of iterations $d_\text{forward}$, yielding a refined $\rcache^{\agbr{d_\text{forward}}}$ that will be used in the final forward pass all the way up to the last layer.

Note that since the above computation is fully differentiable, gradient computation can be back-propagated with recurrence depth of $\dforward$ as well. We allow a separate hyperparameter $\dbackward$ controlling the backward depth analogous to truncated back propagation through time (TBPTT) in classical RNN training \citep{williams2013gradient}.

Throughout the paper, we use $\dforward=16$ and $\dbackward=4$ for experiments unless explicitly stated otherwise. We justify the empirical choices of such parameters as balance of approximation quality and training efficiency (see more analysis in \cref{sec:apdx-bfad_analysis}). We would also want to highlight that while the middle layer recurrence does increase the compute budget necessary for training, the inference cost remains nearly identical to that of the standard Transformer during auto-regressive generation. The only additional overhead comes from the constant compute per step added by the fusion module.

%% file: Sections/expressivity_seperation.tex
\section{Best of Both Worlds: State Tracking and Retrieval}
\label{sec:expressivity_seperation}

Before testing \ours on general natural language modeling, we first isolate its core inductive bias on a synthetic benchmark.
We use $S_5$-Retrieval, a task constructed to sharply separate \ours with (i) standard Transformers, which excel at in-context retrieval but struggle with state tracking at small depth, and (ii) recurrent models (e.g., RNNs/LSTMs), which naturally support state tracking but bottleneck retrieval through a fixed state size.

\subsection{The $S_5$-Retrieval Task}
The $S_5$ state-tracking task, first described by \cite{liu2023transformers}, is a sequence to sequence modeling task where the input is an ordered random sequence of $N$ elements $\rbr{a_1,a_2,\dots, a_N}$ drawn from $S_5$ (the permutation group on $5$ elements, with a cardinality of $120$) and the output is the cumulative composition of the input sequence $\srbr{a_1,a_1\circ a_2,\dots, \Pi_{i=1}^Na_i}$. Here the $i$-th element is considered as the $i$-th state, which transits into the $i+1$-th state by applying $a_{i+1}$ to itself.

Leveraging circuit complexity arguments, \cite{merrill2024illusionstatestatespacemodels} showed that Transformers require $\Omega\rbr{\log N}$ layers to exactly solve the $S_5$ state-tracking, and the known shallowest learnable solution in transformers is via parallel associative scan which requires $\lceil\log_2 N\rceil$ layers \citep{li2025how}. On the other hand, RNNs can solve the task with just constant number of layers \citep{merrill2024illusionstatestatespacemodels}, since its circuit depth grows along the temporal dimension. However, classical recurrent architectures fail in key-value retrieval tasks as the associations must be compressed into a fixed-dimensional state.

To stress test a model's capability of composing both state tracking and in-context retrieval, we define the $S_5$-Retrieval Task as follows (see sample data in \cref{tab:task_example}):

Let $\alphabet$ be an alphabet, and let $\mathcal{D}:S_5\to \alphabet^K$ be a random mapping from the set $S_5$ to length-$K$ strings from the alphabet.
The input is a serialization of the dictionary $\{(a,\mathcal{D}(a)) : a\in S_5\}$ followed by a delimiter and the action sequence $(a_1,a_2,\ldots,a_N)$ (appropriately padded to match token count).
The target output is the step-wise state-tracking results interleaved with retrieval results based on the state:
\[
\rbr{a_1,\quad \mathcal{D}(a_1),\quad a_1\circ a_2,\quad \mathcal{D}(a_1\circ a_2),\quad\ldots,\quad \Pi_{i=1}^Na_i,\quad \mathcal{D}\rbr{\Pi_{i=1}^Na_i}}.
\]

\[
\rbr{a_1,\quad a_2,\quad a_3,\quad \ldots,\quad a_N}\Rightarrow
\rbr{a_1,\quad a_1\circ a_2,\quad a_1 \circ a_2 \circ a_3, \ldots,\quad \Pi_{i=1}^Na_i}
\]

\subsection{Experiments: Empirical Separation between \ours vs. Transformers / RNNs}
\label{sec:s5-result}

We train the $S_5$-Retrieval task on (i) a small 4-layer LSTM model, (ii) a 4-layer, 6-head Llama \citep{touvron2023llama} transformer model, and (iii) a \ours model built on top of the small Llama model with $\lstart=0$ and $\lend=4$, all with matching number of parameters. 

We use $K=4$ and uniformly sample the number of states $N$ from $\cbr{1,2,\dots, 32}$ within the training set. We randomly resample the dictionary $\mathcal{D}$ for each sequence to ensure the necessity of doing purely in-context retrieval.
We train the models with learning rate $1\times 10^{-3}$ and batchsize of 64 ($150$k steps for \ours and $400$k steps for the baselines), and evaluate on a held-out set of input-output pairs with $N$ ranging from $1$ to $48$.

\begin{figure}[H]
    \centering
    \includegraphics[width=0.98\linewidth]{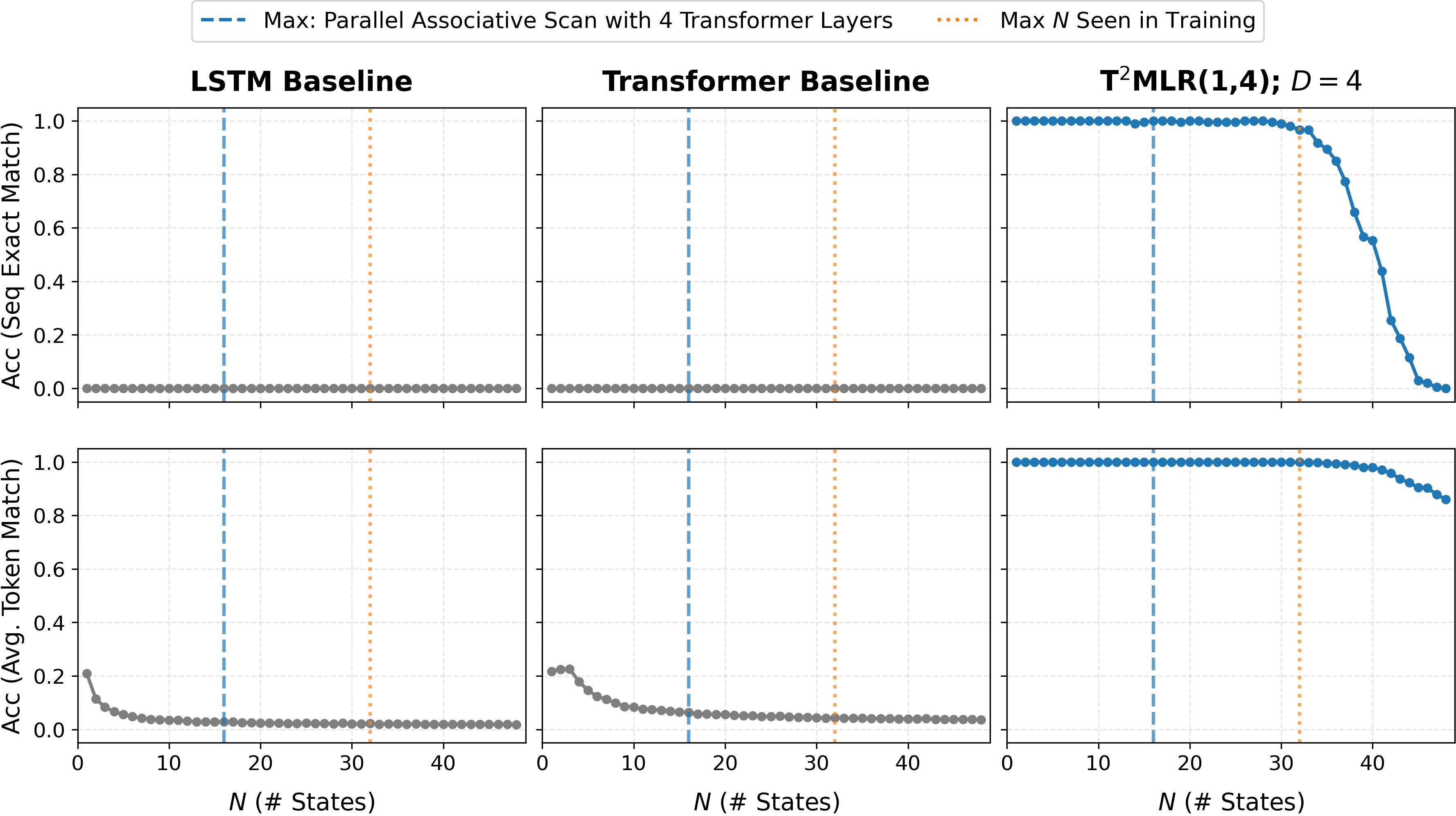}
    \caption{Performance comparison between LSTM, Transformer, and \ours on the $S_5$-Retrieval task after extended training ($150$k steps for \ours, $400$k steps for the baselines). $x$-axis denotes the test length; top row reports sequence exact-match accuracy and bottom row reports average per-token accuracy. The blue dashed line marks the maximum number of states a standard 4-layer Transformer can track using the parallel associative-scan circuit \citep{li2025how}; the orange dotted line marks the maximum number of states seen during training. The exact-match metric is far more sensitive to compounding errors at long lengths; \ours attains near-perfect in-distribution exact match while retaining non-trivial token accuracy at out-of-distribution lengths, whereas the baselines fail to learn the task.}
    \label{fig:s5_token_acc}
\end{figure}

In \cref{fig:s5_token_acc}, we compare \ours against recurrent and Transformer baselines on the $S_5$-Retrieval task, reporting both sequence exact-match accuracy (top row) and average per-token accuracy (bottom row). \ours substantially outperforms both baselines across input lengths, maintaining high exact-match accuracy over a wide range of sequence lengths.  This matches the intended inductive bias of \ours: the recurrent pathway supports continuous propagation of the evolving latent state, while the Transformer backbone preserves access to the input sequence for retrieval.




The sequence exact-match metric (top row of \cref{fig:s5_token_acc}) is highly sensitive to compounding errors at long sequence lengths (each sequence contains up to $48$ states $\times\,5$ tokens per state $=240$ tokens), so it can fall to zero for the baselines even when they are partially correct. The average per-token accuracy (bottom row) gives a finer-grained picture. \ours attains near-perfect in-distribution exact match while sustaining non-trivial token accuracy at higher out-of-distribution lengths, whereas the LSTM and Transformer baselines retain only modest token accuracy and collapse on exact match. The blue dashed line marks the maximum number of states a standard 4-layer Transformer can track using the parallel associative-scan circuit \citep{li2025how}, and the orange dotted line marks the maximum number of states seen during training.

%% file: Sections/experiments.tex
\section{Experiments}

In this section, we provide empirical results of pretraining / finetuning a small \ours model on natural language data as well as synthetic reasoning tasks.
Across a diverse set of tasks we considered, \ours achieves notable gains over the parameter-matched transformer baseline, and middle-layer recurrence generally yields larger gains compared to full-model recurrence. Section~\ref{sec:mechanistic-analysis} contains further mechanistic experiments on \ours, in particular future token prediction.


\subsection{Pretraining \ours}
\label{sec:exp-pretraining}
We first test \ours in standard autoregressive language modeling. We set the baseline as SmolLM2-135M \citep{allal2025smollm2smolgoesbig}, a lightweight LLaMA-like decoder-only Transformer architecture. We construct \ours variants on top of the same backbone, parameterized by recurrence boundaries $(\lstart,\lend)$. We denote the recurrence depth by $D=\lend-\lstart+1$, and refer to each model as \ours$(\lstart,\lend)$ (optionally annotated with $D$).

The representation fusion module introduces additional parameters. To ensure fair comparison, we adjust the baseline hidden size from 576 to 584 so that all models have approximately identical parameter counts (136.4M), with the baseline being slightly larger and thus conservative.
We train all models for one epoch on the official 10B-token FineWeb-Edu subset \citep{penedo2024finewebdatasetsdecantingweb} using identical optimization hyperparameters (\cref{sec:apdx-training_details}), and evaluate zero-shot downstream performance using lm-eval-harness \citep{eval-harness}.

Since most related works on continuous latent reasoning do not easily generalize to pretraining settings due to the lack of sequence-parallelism support, we limit our comparison to the standard transformer baseline. Unless otherwise noted, our comparisons are matched on parameter count and inference-time compute. \ours incurs additional training overhead due to the Jacobi-style approximation described in \cref{sec:bfa}; we report a training-compute-matched comparison in \cref{tab:compute_matched} and discuss this trade-off in \cref{sec:discussion}.

\begin{table}[H]
\centering
\begin{adjustbox}{width=\textwidth}
\begin{tabular}{l|ccccccc|c}
\toprule
\toprule
Model/Config & ARC-C & ARC-E & HS & OBQA & PIQA & SciQ & WG & Average \\
Metric & acc\_n $\uparrow$ & acc\_n $\uparrow$ & acc\_n $\uparrow$ & acc\_n $\uparrow$ & acc\_n $\uparrow$ & acc\_n $\uparrow$ & acc $\uparrow$ & - $\uparrow$ \\
\midrule
Baseline & \textbf{24.74} & 44.28 & 29.81 & 30.20 & 61.53 & 60.80 & 48.46 & 42.83 \\
\ours(1,30); $D{=}30$ & 24.49 & 43.48 & 29.98 & 30.00 & 60.72 & 62.60 & \textbf{52.25} & 43.36 \\
\ours(5,26); $D{=}22$ &  23.98 & \textbf{46.13} & \textbf{30.75} & 29.80 & 60.77 & 62.80 & 51.85 & 43.73 \\
\ours(9,22); $D{=}14$ & 24.15 & 45.24 & 30.40 & 29.20 & 61.15 & \textbf{66.40} & 51.54 & 44.01 \\
\ours(13,18); $D{=}6$  & 24.23 & 45.50 & 29.95 & \textbf{31.20} & \textbf{61.70} & 64.20 & 52.17 & \textbf{44.14} \\
\ours(15,16); $D{=}2$ & 24.40 & 45.83 & 30.11 & 29.20 & 59.63 & 60.20 & 50.83 & 42.89 \\
\bottomrule
\end{tabular}
\end{adjustbox}
\caption{Zero-shot downstream evaluation of 135M \ours variants with different recurrence boundaries $(\lstart,\lend)$, pretrained on 10B FineWeb-Edu tokens. All results are obtained using lm-eval-harness \citep{eval-harness}; we report normalized accuracy (acc\_n) whenever available. Abbreviations: ARC-C/E = ARC-Challenge/Easy, HS = HellaSwag, OBQA = OpenBookQA, WG = Winogrande.}
\label{tab:eval_135m}
\end{table}

\begin{minipage}[t]{0.68\textwidth}
    As shown in \cref{tab:eval_135m}, \ours consistently matches or improves upon the parameter-matched Transformer baseline on most downstream benchmarks.
Across recurrence configurations,  \ours(13,18) ($D{=}6$) achieves the highest average score while only looping over 20\% of the layers. However, full recurrence ($D{=}30$), as used by all the previous latent reasoning works, generally performs weaker than middle layer recurrence.

\vspace{0.1in}
In \cref{fig:loss_vs_lstart}, we show the training loss for different variants of \ours. Most configurations attain lower evaluation loss compared to the baseline except for $\lstart=1$ (full recurrence with $D=30$) and $\lstart=14$ (only recurring on $D=2$ layers). Apart from these two exceptions, we note that the evaluation loss generally improves with increasing number of recurrent layers, yielding different trend as in downstream. This suggest \emph{distinct implicit bias of middle layers} on reasoning that might not be captured by perplexity metrics and resembles the observation in \cite{saunshi2024inductive} for middle layer stacking.
\end{minipage}%
\hfill 
\begin{minipage}[t]{0.29\textwidth}
\begin{figure}[H]
    \centering
    \vspace{-0.2in}
    \includegraphics[width=0.95\linewidth]{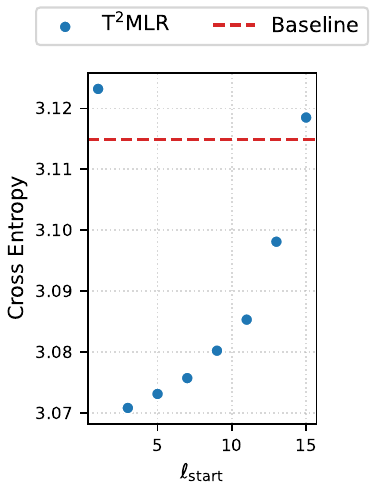}
    \vspace{-0.05in}
    \caption{Validation cross-entropy loss vs $\lstart$ for language modeling pretraining runs at 16k steps.}
    \label{fig:loss_vs_lstart}
\end{figure}
\end{minipage}

We further find that these improvements are not specific to the 135M from-scratch setting: they persist and grow when scaling to 361M and 1B parameters and when extending pretraining to 50B tokens, and a fixed-width ablation confirms that middle-layer placement remains optimal at fixed recurrence depth. We defer the full scaling results and recurrence-location ablation to \cref{sec:apdx-scaling}.

\subsection{Comparison to Looped and Latent-Recurrent Baselines}
\label{sec:exp-looped}

Our main controlled comparison is against parameter- and data-matched Transformers, which isolates the effect of the temporal middle-layer pathway. To further situate \ours among architectures that also add latent or recurrent computation without replacing the attention sequence mixer, we compare against three closely related baselines at 135M scale, all matched on parameter count and trained on the same 10B FineWeb-Edu tokens: a $2\times$ pause-token model \citep{merchant2023pause}, a $2\times$ full-looped Transformer, and a $3\times$ middle-looped Transformer (looping layers 9--22). We do not directly compare against COCONUT-style embedding-level latent recurrence \citep{hao2024coconut}, as those methods lack a scalable dense teacher-forcing pretraining mechanism.

As shown in \cref{tab:looped_baselines}, \ours attains the best average performance among these baselines. Crucially, while \ours incurs extra forward iterations only during \emph{training}, the looped and pause-token baselines incur additional \emph{inference} cost at every decoding step: the pause-token variant scales quadratically with the number of pauses, and the looped variants multiply per-token compute by the number of loops. \ours therefore matches or exceeds these baselines while retaining standard autoregressive inference cost (\cref{sec:apdx-inference-overhead}).

\begin{table}[H]
\centering
\begin{adjustbox}{width=\textwidth}
\begin{tabular}{l|ccccccc|c}
\toprule
\toprule
Model/Config & ARC-C & ARC-E & HS & OBQA & PIQA & SciQ & WG & Average \\
Metric & acc\_n $\uparrow$ & acc\_n $\uparrow$ & acc\_n $\uparrow$ & acc\_n $\uparrow$ & acc\_n $\uparrow$ & acc\_n $\uparrow$ & acc $\uparrow$ & - $\uparrow$ \\
\midrule
Transformer Baseline & 24.74 & 44.28 & 29.81 & 30.20 & 61.53 & 60.80 & 48.46 & 42.83 \\
\ours(9,22) & 24.15 & 45.24 & \textbf{30.40} & 29.20 & 61.15 & \textbf{66.40} & 51.54 & 44.01 \\
\ours(13,18) & 24.23 & \textbf{45.50} & 29.95 & 31.20 & \textbf{61.70} & 64.20 & \textbf{52.17} & \textbf{44.14} \\
Pause-token $\times 2$ & 24.74 & 44.57 & 29.51 & \textbf{31.60} & 60.23 & 61.90 & 50.59 & 43.31 \\
Full-looped $\times 2$ & \textbf{24.83} & 44.95 & 29.84 & 30.80 & 60.23 & 60.70 & 49.57 & 42.99 \\
Middle-looped $\times 3$ & 23.04 & 45.16 & 30.12 & 30.20 & 60.17 & 59.80 & 50.28 & 42.68 \\
\bottomrule
\end{tabular}
\end{adjustbox}
\caption{Comparison against looped and latent-recurrent baselines at 135M (10B FineWeb-Edu tokens), all matched on parameter count. \ours achieves the best average while, unlike the looped and pause-token baselines, adding no per-token inference overhead.}
\label{tab:looped_baselines}
\end{table}


\subsection{Finetuning \ours on Reasoning Downstream Tasks} \label{sec:multi-hop-datasets}

While perplexity reflects average next-token prediction, it can understate changes in the model's internal computation.
To directly test whether temporal middle-layer recurrence improves latent reasoning, we finetune the pretrained checkpoints from \cref{sec:exp-pretraining} on the following suite of multi-hop reasoning and grade-school math tasks:

\textbf{Variable assignment:}\quad
We take the variable assignment task from the set of "reasoning-primitives" task proposed by \cite{saunshi2024inductive}. In this task, the input is a series of variable assignments, and the model needs to output the value of a query variable. Since the baseline models already saturates the original dataset, we increase the difficulty of the task from depth$=2$ to depth$=5$.

\textbf{ProsQA-Hard:}\quad
We evaluate on the ProsQA task from \cite{hao2024coconut}. The input is a directed-acyclic-graph, and a query asks a binary question for the connectivity between a starting and two choices of ending nodes. The output steps through a path that connects the starting and end nodes. Since the original ProsQA data is saturated by the baseline model, we increase the average number of nodes to 60 and average length of the path to 8.

\textbf{HotPotQA-Simple:}\quad
HotpotQA is a question answering dataset featuring natural, multi-hop questions, with annotated supporting facts \citep{yang2018hotpotqa}. We use a teacher model (GPT-4o-mini \citep{openai2024gpt4ocard}) to generate natural language chain-of-thought SFT data. Since the medium and hard level subsets are too challenging at the model scale we are testing, we present the results for the easy subset.

\textbf{GSM-Aug (Symbolic / Natural Language):}\quad To check whether the \ours improves arithmetic-heavy reasoning, we finetune the pre-trained checkpoints on GSM-8K-style grade-school math problems. In particular, we use synthetic GSM style training data from \cite{deng2023implicitcot}, which contains a symbolic variant only containing the arithemetic computation involved, and a natural language version more analogous to natural CoT reasoning.

\begin{figure}[H]
  \centering
  \includegraphics[width=\textwidth]{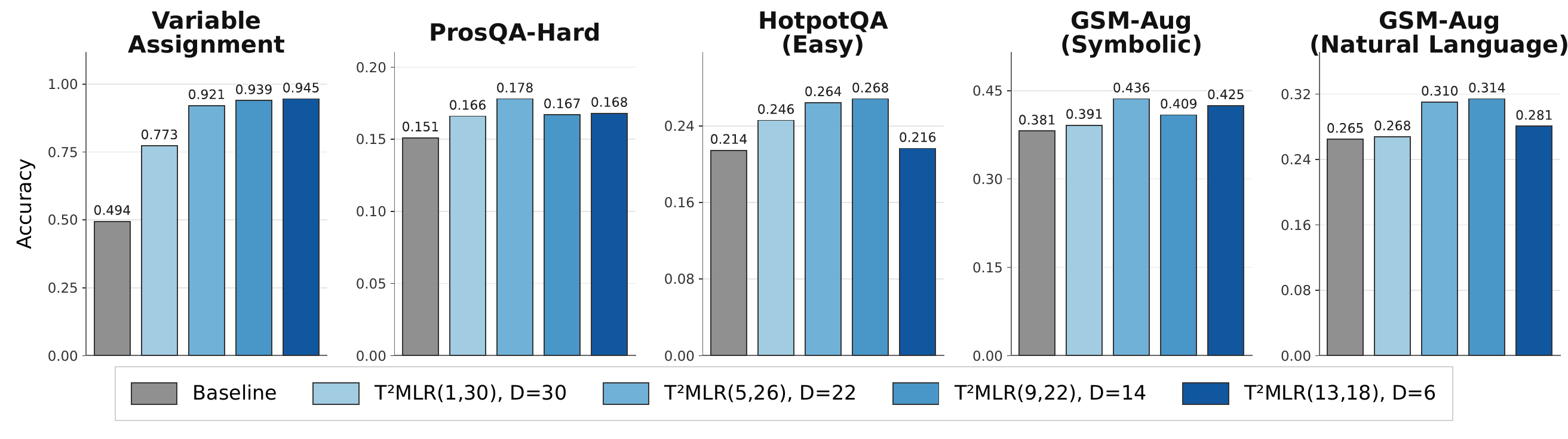}
  \caption{Fine-tuning performance comparison between variants of \ours and transformer baseline on multi-hop reasoning and grade-school math tasks. We report pass@1 accuracy for all experiments. In all five tasks, \ours outperforms parameter and data-matched transformer baselines. Notably, we see the middle-layer looping variants ($D=22,14,6$) achieving higher performance than full looping ($D=30$).}
  \label{fig:finetuning-multihop}
\end{figure}

We report the downstream results in \cref{fig:finetuning-multihop}. On both synthetic and real world reasoning tasks, \ours with middle-layer recurrence shows better performance compared to the parameter and data-matched baseline. Meanwhile, the full layer recurrence model shows smaller improvements. This further highlights the importance of leveraging middle layer recurrence to boost latent reasoning capabilities.

\subsection{Retrofitting a Pretrained Transformer into \ours}
\label{sec:exp-retrofit}

A key question for practical adoption is whether \ours must be trained from scratch, or whether the recurrent pathway can be grafted onto an existing pretrained model. This is our largest-scale experiment, and the most direct test of real-world applicability: we retrofit a pretrained \emph{instruction-tuned} SmolLM2-1.7B-Instruct model (32 layers) by inserting the recurrent fusion pathway (\ours(5,28)) and continuing finetuning for one epoch on the OpenMathReasoning dataset \citep{openmathreasoning2024}, comparing against the identical model finetuned \emph{without} the recurrent pathway.

As shown in \cref{tab:retrofit}, retrofitting improves GSM8K from $35.78$ to $39.88$ ($+11.5\%$ relative) and MATH500 from $12.80$ to $18.00$ ($+40.6\%$ relative). In other words, simply adding the temporal middle-layer pathway to an off-the-shelf pretrained model and briefly finetuning delivers substantial reasoning gains, \emph{without} any pretraining from scratch. This makes \ours markedly easier to adopt: existing pretrained Transformers can be upgraded with the recurrent pathway at a fraction of the cost of pretraining, while retaining the near-identical inference cost of the base model.

\begin{table}[H]
\centering
\begin{tabular}{l|cc}
\toprule
\toprule
Model & GSM8K & MATH500 \\
\midrule
Baseline (SmolLM2-1.7B-Instruct) & 35.78 & 12.80 \\
\ours(5,28) & \textbf{39.88 {\scriptsize (+11.5\%)}} & \textbf{18.00 {\scriptsize (+40.6\%)}} \\
\bottomrule
\end{tabular}
\caption{Retrofitting a pretrained SmolLM2-1.7B-Instruct model into \ours via continued finetuning on OpenMathReasoning \citep{openmathreasoning2024}. Adding the recurrent fusion pathway improves both GSM8K and MATH500.}
\label{tab:retrofit}
\end{table}

%% file: Sections/mechanistic_interpretation.tex


%% file: Sections/related_works.tex
\section{Related Works}

\textbf{Latent reasoning}\quad
Recent latent-reasoning methods relax the token-space information bottleneck by shifting part of the reasoning process from explicit text into continuous hidden representations. Coconut \citep{hao2024coconut} and CODI \citep{shen2025codi} internalize chain-of-thought into latent states, replacing portions of textual reasoning with hidden-state computation. Related works replace discrete reasoning tokens with soft or continuous alternatives, such as mixed latent/text training and continuous token mixtures in embedding space \citep{su2025tokenassorted,tack2025cocomix,zhang2025soft,butt2025softtokenshardtruths,zhuang2025mixture,tang2026multiplexthinkingreasoningtokenwise}. Across these approaches, the common theme is to reduce reliance on verbose discrete reasoning traces by allowing computation to proceed in continuous space.

\ours differs the most from existing works by relaxing the bottleneck through \emph{middle-layer recurrence}. This preserves intermediate reasoning states in the part of the network where abstract computation occurs. In addition, \ours is trainable with standard supervised fine-tuning under dense token-level supervision, and allows BPTT to optimize the recurrent pathway across tokens. The closest prior work is HRPO \citep{yue2025hybridlatentreasoningreinforcement}. However their method feeds latent information back through the embedding interface and is trained in an RL setting without gradient flow through past tokens. \Cref{tab:latent_reasoning_comparison} provides an additional comparison between \ours and other latent reasoning works.

\textbf{Looped Transformers}\quad
Looped transformers reuse the same block multiple times to increase effective reasoning depth \citep{dehghani2018universal,giannou2023looped}. Notably, \cite{geiping2025recurrentdepth} and \cite{zhu2025ouro} leverage KV caches from later loops in earlier loops, representing another form of shortcutting deeper representations to shallower layers. Unlike \ours, which adds at most $\sim$8\% per-token inference overhead (\cref{sec:apdx-inference-overhead}), looped architectures incur increased inference costs that scale with the number of loops.

\textbf{State-space, hybrid, and recurrent-memory models}\quad
A related body of work introduces recurrence along a \emph{different architectural axis}: state-space and hybrid models (e.g., Mamba \citep{gu2023mamba}, RetNet \citep{sun2023retnet}, Griffin \citep{de2024griffin}) replace or interleave attention with recurrent/linear-time sequence mixers, typically motivated by long-context efficiency, while recurrent-memory variants such as Transformer-XL \citep{dai2019transformerxl} carry memory across segments beyond the context window. In contrast, \ours keeps the Transformer backbone and standard attention intact and instead adds a \emph{deep-to-shallow} recurrent residual pathway across decoding steps, targeting reasoning within a shorter context window rather than long-context memory.
Due to the limited space, we defer the more extended discussion around recurrent memory and state-space models / linear attention works to Appendix \cref{app:related_works}.

%% file: Sections/discussion.tex
\section{Discussions, Limitations, and Future work}
\label{sec:discussion}

\textbf{Training-time overhead}\quad
The main limitation of \ours is training cost. Because our batch approximate forward applies multiple Jacobi-style refinement steps through the recurrent middle block, training is slower than a parameter-matched Transformer baseline. In our pretraining setting, a moderate approximation depth is already sufficient, but this still leads to roughly a $2$--$4\times$ wall-clock overhead depending on the size of the recurrent block $\lend - \lstart$. We make this trade-off explicit in \cref{tab:compute_matched}: under a \emph{training-wall-clock-matched} budget, a longer-trained Transformer can surpass the 135M \ours on zero-shot NLP. Our parameter-, data-, and inference-compute-matched comparisons should therefore be read as isolating the \emph{architectural and inference-side} contribution of temporal middle-layer recurrence---which adds at most $\sim$8\% per-token inference overhead (\cref{sec:apdx-inference-overhead})---rather than as a claim of a training-compute win. A key direction for future work is to reduce this training cost further by improving the temporal mixing mechanism or the approximation scheme. In particular, it may be possible to preserve the benefits of middle-layer recurrence with fewer refinement steps, or to reuse exact recurrent states more directly in settings such as on-policy RL, where those states are naturally produced during rollout generation.

\textbf{Scale and Transfer}\quad
Our main experiments focus on relatively small models, but we find that the gains from middle-layer recurrence persist as we scale to 361M and 1B parameters and extend pretraining to 50B tokens (\cref{tab:eval_scale,tab:eval_50b}), with the relative improvements on reasoning-oriented tasks growing at larger scale. We also show that \ours need not be trained from scratch: retrofitting a pretrained SmolLM2-1.7B-Instruct model by inserting the recurrent fusion module and continuing finetuning yields substantial gains on GSM8K and MATH500 (\cref{tab:retrofit}), making the architecture substantially easier to adopt in practice. Scaling further to larger models and more demanding reasoning benchmarks, as well as multi-seed variance estimates, remain important directions for future work.

\textbf{Architecture Design and Mechanistic Understanding}\quad
We believe our recurrent fusion module as well as the general recurrent pathway can be further improved. Understanding which fusion parameterizations and recurrence boundaries work best may lead to stronger and more efficient variants. It is also an interesting future direction to understand the information propagated through the recurrent pathways via a mechanistic lens.

\section{Conclusion}

We introduced \ours, a Transformer architecture that routes recurrence through the middle layers. Across pretraining and downstream reasoning tasks, \ours consistently outperforms parameter- and data-matched Transformer baselines, with localized middle-layer recurrence often outperforming full-layer recurrence. Taken together, these results suggest that the value of recurrence in Transformers depends not only on enabling iterative latent computation, but on introducing it at the right depth.

We hope this work serves as a starting point for a broader investigation of \emph{where} recurrence should live in latent-reasoning architectures, and whether reasoning gains can be unlocked by placing recurrent computation where abstract processing is most active.  




%% file: Appendix/Sections/appendix.tex
\newpage
\input{Appendix/Sections/related_works_full}
\newpage
\input{Appendix/Sections/bfa}
\newpage
\input{Appendix/Sections/additional_results}
\newpage
\input{Appendix/Sections/future_token_prediction}

\newpage

\input{Appendix/Sections/training_analysis}
\newpage
\input{Appendix/Sections/experiment_setup}

%% file: Appendix/Sections/related_works_full.tex
\section{Extended Related Works}
\label{app:related_works}

\paragraph{Latent Reasoning}
Many recent works study or exploit \emph{latent computation} in transformers to improve reasoning without relying on long explicit chain-of-thought (CoT).


On the modeling side, recent work has explored increasing reasoning capacity without relying on long explicit chain-of-thought traces, by shifting computation into latent or continuous representations. One line of work focuses on \emph{internalizing} chain-of-thought into hidden states, via distillation or progressively removing explicit intermediate steps while preserving final-answer performance \citep{deng2023implicitcot,deng2024explicit2implicit,hao2024coconut,shen2025codi}. A second line explicitly allocates additional \emph{inference-time computation} to latent processes, such as iterative latent refinement, token-wise branch-and-merge reasoning, or latent communication in multi-agent settings \citep{kong2025ltm,fu2025thinkathard,tang2026multiplexthinkingreasoningtokenwise,zou2025latentcollaborationmultiagentsystems}. Finally, several approaches replace discrete reasoning tokens with \emph{soft or continuous} alternatives, including mixed latent/text token training and continuous mixtures in embedding space, which can shorten or eliminate verbose reasoning traces \citep{su2025tokenassorted,tack2025cocomix,zhang2025soft,butt2025softtokenshardtruths,zhuang2025mixture}. Additionally, \cite{zhu2025superposition} finds latent COT offers theoretical benefits on graph tasks.


Our work is most similar to \cite{yue2025hybridlatentreasoningreinforcement}, which uses latent information by projecting the last-token latent state back to token space before the first layer. They use the resulting architecture for RL training, showing improvements on mathematical reasoning benchmarks. Crucially, in \cite{yue2025hybridlatentreasoningreinforcement} gradients do not flow to the past tokens, while in our work, we use supervised training and back-propagate through time. Furthermore, we emphasize \emph{middle-layer} recurrence, avoiding constraints that force recurrent information to live in the embedding space.


\begin{table}[H]
\centering
\newcommand{\celltab}[1]{%
  {\renewcommand{\arraystretch}{0.9}%
  \begin{tabular}[c]{@{}l@{}}#1\end{tabular}}%
}
\renewcommand{\arraystretch}{1.2}
\small
\begin{adjustbox}{width=\columnwidth,center}
\begin{tabular}{l|l|l|l|c}
\toprule
Model & Recurrent Information & Injection Location & Fusion Module & BPTT \\
\midrule

Transformer
& Discrete token
& Input embedding
& --
& N/A \\

Coconut
& Last-layer hidden state
& Post-embedding
& Identity
& $\checkmark$ \\

CODI
& Last-layer hidden state
& Post-embedding
& 2-layer MLP
& $\checkmark$ \\

Multiplex-Thinking
& Soft token mixture
& Post-embedding
& Identity
& $\times$ \\

HRPO
& \celltab{Soft token mixture \\ + discrete token}
& \celltab{Post-embedding (soft token) \\ + Input embedding (discrete token)}
& Gated fusion
& $\times$ \\

\midrule

\ours (ours)
& \celltab{
Middle-layer hidden state \\
+ Discrete token
}
& \celltab{
Earlier middle layer (hidden state) \\
 + Input embedding (discrete token)
}
& Gated fusion
& $\checkmark$ \\

\bottomrule
\end{tabular}
\end{adjustbox}
\caption{Comparison with related architectures. Standard transformers communicate across time only through discrete tokens. Prior recurrent or latent-reasoning variants additionally propagate either last-layer hidden states or soft token mixtures, usually through input-level or post-embedding pathways. \ours adds a recurrent middle-layer hidden-state pathway, fused into an earlier middle layer of the next token, while preserving the ordinary discrete token pathway and supporting end-to-end supervised training with a constant-depth BPTT.}
\label{tab:latent_reasoning_comparison}
\end{table}

\paragraph{Looped Transformers}

Looped (recurrent-depth) Transformers trade new parameters for repeated computation by applying the same Transformer block multiple times. Universal Transformers introduce this idea explicitly as recurrence over depth, optionally with adaptive halting \cite{dehghani2018universal}. ALBERT popularizes cross-layer parameter sharing in Transformers to reduce parameter count \cite{lan2019albert}. On the expressivity side, Giannou et al.\ show that looping expands the expressive power of Transformers \cite{giannou2023looped}. \cite{saunshi2025latent} provide theoretical evidence that looping can approximate the benefits of much deeper Transformers, yielding large effective depth with fewer parameters.

\cite{geiping2025recurrentdepth} demonstrate empirically that increasing test-time compute via recurrent depth improves reasoning. Most recently, ByteDance's Ouro \citep{zhu2025ouro} looped LMs scale this design to billion-parameter models, reporting performance that rivals much larger-parameter Transformers. It is worth noting that both \cite{geiping2025recurrentdepth} and \cite{zhu2025ouro} introduced leveraging the KV caches of the last loops when computing the forward pass for the shallower loops. This represents another form of creating a shortcut for utilizing more refined deeper representations at shallower layers.

\paragraph{Recurrent memory for long-context modeling.}
A line of work uses recurrence primarily as a \emph{memory mechanism} to improve long-context performance, rather than to change per-token computation. Transformer-XL reuses hidden states across segments to extend effective context \citep{dai2019transformerxl}, while Compressive Transformer retains older history via a compressed memory \citep{rae2019compressive}. Recurrent Memory Transformer passes information across segments using dedicated memory tokens \citep{bulatov2022rmt}. More recently, Titans \citep{behrouz2024titans} augments standard attention over a fixed-length current context (motivated by attention's quadratic cost) with a neural long-term memory module that stores and retrieves information from much longer histories.

\paragraph{State space and hybrid models.}
State space models offer subquadratic alternatives to attention by maintaining and updating a compact state over the sequence, with representative examples including S4 \citep{gu2021s4}, Hyena \citep{poli2023hyena}, and Mamba \citep{gu2023mamba}. Some hybrid architectures interleave attention with recurrent/SSM-style blocks to trade off content-based retrieval and efficiency, e.g., RetNet \citep{sun2023retnet}, Griffin \citep{de2024griffin}, and DeltaNet \citep{yang2024latently,yang2024parallelizing}. In contrast, we do not introduce a new recurrent/SSM layer or replace attention; instead, we add a token-to-token recurrent connection that injects a previous token's later-layer residual stream into an earlier-layer residual stream of the current token.

%% file: Appendix/Sections/bfa.tex
\section{Additional Information on Training \ours}
\label{sec:apdx-bfa}



\subsection{Temporal-Parallel Approximated Training}

Here we formally describe the fixed-point iteration algorithm used to train \ours, which has been informally introduced in \cref{sec:bfa}.

Let $\mH^{(0)}\in\R^{T \times d}$ denote the input embedding for a sequence of $T$ tokens, and with slight abuse of notation let $\Fl{\ell_1:\ell_2}: \R^{T \times d}\to \R^{T \times d}$ denote the mapping on embedding sequences implemented by layers between $\ell_1$ and $\ell_2$. After doing a standard forward pass (assuming no recurrent cache exists), we store $\mH^{(\lstart-1)}:=\Fl{1:\lstart-1}(\mH^{(0)})$ as the exact input into layer $\lstart$. This also corresponds to the exact input for the sequence if recurrence has been done sequentially.

During the same forward pass, we set the right-shifted version of $\mH^{(\lend)}_{\langle 0\rangle}:=\Fl{1:\lend}(\mH^{(0)})$ as the first iterate of the approximated recurrent cache. Formally, denote $\rcache_{\agbr{1}}:=\mathrm{ShiftRight}(\mH^{(\lend)}_{\langle 0\rangle})$. Here, we use subscripts in angle brackets to denote the index of iteration for the recurrent cache and the subsequent representation induced by that cache.

With this approximated cache, we merge with the cached $\mH^{(\lstart-1)}$ through the fusion module $\Fuse(\cdot,\cdot)$. Then we pass the fused representation through the middle layers, getting \begin{equation}
     \mH^{(\lend)}_{\langle 1\rangle} := \Fl{\lstart:\lend}\!\Big(\Fuse\rbr{\mH^{(\lstart-1)}_{\langle 0\rangle},\, \rcache_{\langle 1\rangle}}\Big).
\end{equation}
This gives us the post-$\lend$ representation conditioned on the first iterate of the cache $\rcache_{\langle 1\rangle}$. We then combine the right-shifted $\rcache_{\langle 1\rangle}$ with right shifted $\mH^{(\lend)}_{\langle 1\rangle}$ following the temporal cache residual rule to yield the next iterate of cache $\rcache_{\langle 2\rangle}$. We continue this iterative update for $d_{\text{forward}}$ time to $\rcache_{\langle D_{\text{forward}}\rangle}$, and use it as the recurrent cache for the final forward pass.

To avoid too deep of a gradient backward pass, we also allow selective detachment analogous to TBPTT in standard RNN training \citep{williams2013gradient}.
We present the full temporal-parallel approximated forward algorithm in \cref{alg:bfa_tmlr_minimal}, where we adopt a temporal residual connection for the cache as described in \cref{sec:gating}.

With no temporal cache residual, line 8 in \cref{alg:bfa_tmlr_minimal} is then
    $\rcache_{\langle k\rangle} \leftarrow \mathrm{ShiftRight}\!\left(\mH^{(\lend)}_{\langle k-1\rangle}\right).$

\begin{algorithm}[H]
\caption{Temporal-Parallel Approximated Forward for \ours}
\label{alg:bfa_tmlr_minimal}
\KwIn{$D_{\textit{forward}}$, $D_{\textit{backward}}$, input embeddings $\mH^{(0)}\in\R^{T\times d}$}
\KwOut{$\mH^{(L)}_{\langle D_{\textit{forward}}\rangle}\in\R^{T\times d}$}

$\mH^{(\lstart-1)} \leftarrow \Fl{1:\lstart-1}\!\big(\mH^{(0)}\big)$ \tcp*[r]{prefill to $\lstart-1$}

$\mH^{(\lend)}_{\langle 0\rangle} \leftarrow \Fl{\lstart:\lend}\!\big(\mH^{(\lstart-1)}\big)$ \tcp*[r]{single pass to seed cache}
$\rcache_{\langle 1\rangle} \leftarrow \mathrm{ShiftRight}\!\left(\mH^{(\lend)}_{\langle 0\rangle}\right)$ \tcp*[r]{initialize deep cache}

\For{$k=2$ \KwTo $D_{\text{forward}}$}{
    $\mH^{(\lend)}_{\langle k-1\rangle} \leftarrow \Fl{\lstart:\lend}\!\Big(\Fuse\rbr{\mH^{(\lstart-1)}_{\langle 0\rangle},\, \rcache_{\langle k-1\rangle}}\Big)$ \tcp*[r]{recurrent refinement}
    \If{$k > D_{\text{forward}} -D_{\text{backward}}$}{
        $\mH^{(\lend)}_{\langle k-1\rangle} \leftarrow \mathrm{StopGrad}\!\left(\mH^{(\lend)}_{\langle k-1\rangle}\right)$ \tcp*[r]{truncate gradients}
    }
    $\rcache_{\langle k\rangle} \leftarrow \mathrm{Normalize}\rbr{\mathrm{ShiftRight}\!\left(\mH^{(\lend)}_{\langle k-1\rangle} + \rcache_{\langle k-1\rangle}\right)}$ \tcp*[r]{update cache}
}

$\mH^{(L)}_{\langle D_{\text{forward}}\rangle} \leftarrow
\Fl{\lstart:L}\!\Big(\Fuse\rbr{\mH^{(\lstart-1)}_{\langle 0\rangle},\, \rcache_{\langle D_{\text{forward}}\rangle}}\Big)$ \tcp*[r]{forward w/ refined cache}

\Return{$\mH^{(L)}_{\langle D_{\text{forward}}\rangle}$}\;
\end{algorithm}

\subsection{Computation Overhead of Temporal-Parallel Jacobi Training}
\label{sec:apdx-compute-overhead}

In \cref{tab:compute_cost}, we summarize the asymptotic computation cost of \ours.

With $N$ as the sequence length, $L$ as the total number of Transformer layers, $d$ as the hidden width, and $D=\lend-\lstart+1$ as the size of the recurrent middle block. For full-sequence training without KV caching, each layer incurs cost $O(N^2 d + N d^2)$, where the $O(N^2 d)$ term comes from dense self-attention and the $O(N d^2)$ term comes from projections and MLPs. For autoregressive inference with KV caching at context length $N$, each layer costs $O(N d + d^2)$ per generated token.

The training overhead of \ours comes from the $\dforward$ Jacobi-style refinement steps used to approximate the recurrent cache in parallel across token positions. Importantly, these extra passes only re-run the recurrent middle block of size $D$, rather than the full $L$-layer network. As a result, the training cost of \ours is $
O\!\left((L+\dforward D)\,(N^2 d + N d^2)\right),
$
compared with
$
O\!\left(L\,(N^2 d + N d^2)\right)
$
for a vanilla Transformer.

This should be contrasted with full-layer looping approaches, whose cost scales as $O(KL\,(N^2 d + N d^2))$, and middle-block looping approaches, whose cost scales as $O((L+(K-1)D)\,(N^2 d + N d^2))$. Thus, the additional cost of \ours scales only with the recurrent block size $D$ and the Jacobi depth $\dforward$, which is substantially cheaper than repeatedly looping over the entire stack when $D \ll L$.

At inference time, \ours does not require extra serial Jacobi refinement. Once the recurrent cache from the previous token is available, the current token is processed with a single forward pass together with one application of the fusion module. Therefore, under KV caching, the per-token inference complexity remains approximately the same as that of a standard Transformer,
$O\!\left(L\,(N d + d^2)\right)$ up to a small constant-factor overhead from the fusion computation. This distinguishes \ours from looped-Transformer approaches, whose inference cost grows linearly with the number of serial loops.

The same argument is particularly favorable in RL settings. During online rollout generation, the model already computes the exact recurrent latent states sequentially token by token, so for rollout tokens there is no need to perform additional forward-side Jacobi refinement. Instead, one can cache these exact latent representations during decoding and reuse them during policy optimization. Temporal-parallel approximation is only needed in settings such as prefill, teacher-forced training, or replayed trajectories where token-parallel processing is desired.

It is worth noting that in RL-style training, $\dforward$ need not be chosen to improve the approximation quality of recurrent cache themselves. It only needs to be large enough to construct the temporal computation graph needed for BPTT. Therefore one can set $\dforward = \dbackward$, which will significantly reduce the practical overhead relative to supervised pretraining.

\begin{table}[t]
\centering
\small
\setlength{\tabcolsep}{5pt}
\begin{tabular}{lcc}
\toprule
\textbf{Model} &
\textbf{Train (per seq)} &
\textbf{Infer (per token)} \\
\midrule
Vanilla
& $O\!\left(L\,(N^{2}d + N d^{2})\right)$
& $O\!\left(L\,(N d + d^{2})\right)$ \\
Full-loop
& $O\!\left(KL\,(N^{2}d + N d^{2})\right)$
& $O\!\left(KL\,(N d + d^{2})\right)$ \\
Mid-loop
& $O\!\left((L+(K-1)D)\,(N^{2}d + N d^{2})\right)$
& $O\!\left((L+(K-1)D)\,(N d + d^{2})\right)$ \\
\ours
& $O\!\left((L+\dforward D)\,(N^{2}d + N d^{2})\right)$
& $\approx O\!\left(L\,(N d + d^{2})\right)$ \\
\bottomrule
\end{tabular}
\caption{\textbf{Asymptotic compute comparison.}
Here $N$ is the sequence length, $L$ the total number of layers, $d$ the hidden width, and $D=\lend-\lstart+1$ the size of the recurrent middle block. }
\label{tab:compute_cost}
\end{table}

\newpage

\subsection{On the Validity of Approximation Depth}\label{sec:apdx-bfad_analysis}

Due to the existence of the $\mathrm{ShiftRight}$ operation, the $t$-th column of $\rcache$ will converge after $t$-iterations. In this subsection, we show that empirically, we might need way fewer depth for training.

We take a middle checkpoint from the FineWeb-Edu pretraining runs (with $\lstart=5, \lend=26, \dforward=16, \dbackward=4$), sample a random training batch with 2048 tokens per sequence, and compute the gradients under different approximation-depth settings $(\dforward,\dbackward)$ in \cref{alg:bfa_tmlr_minimal}. 
We treat the gradients under the largest setting $(\dforward,\dbackward)=(32,32)$ as an \emph{anchor}.

\begin{figure}[H]
    \centering
    \includegraphics[width=0.8\linewidth]{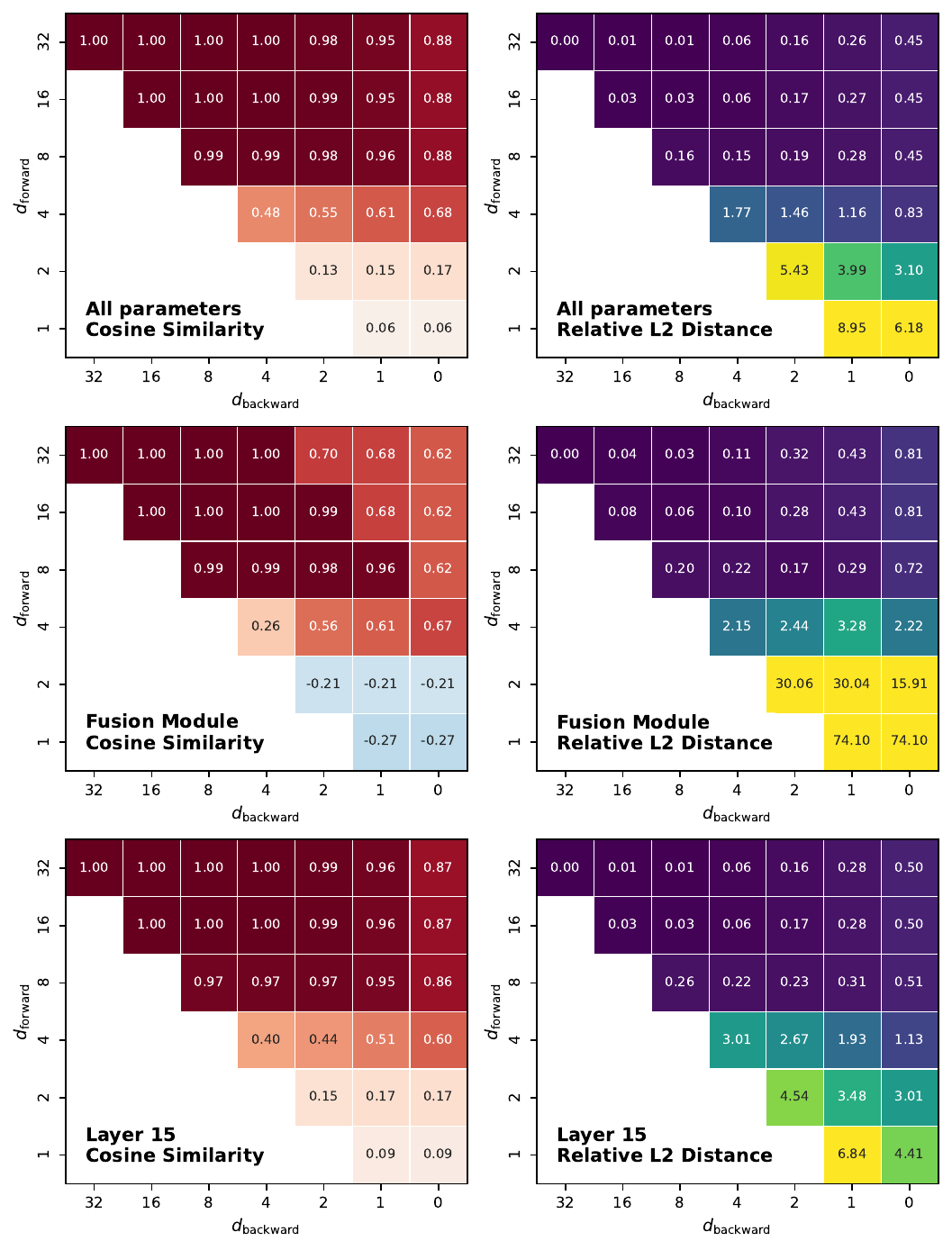}
    \caption{Cosine similarity and relative $\ell_2$ distance between gradients attained under different setup of $(\dforward, \dbackward)$ against setting $\dforward=32, \dbackward=32$.}
    \label{fig:grad_analysis}
\end{figure}

For each $(\dforward,\dbackward)$ setting, we measure the difference between the gradient and the anchor by (i) the cosine similarity and (ii) the $\ell_2$ distance normalized by the norm of the anchor gradients (which we denote by relative $\ell_2$ distance). As shown in \cref{fig:grad_analysis}, setting $\dforward < 8$ and $\dbackward \le 4$ could lead to drastic gradient differences compared to the anchor, while further increasing depth yields diminishing benefits.

\subsection{Jacobi Approximation vs. Exact Recurrence in Training}
\label{sec:apdx-jacobi-vs-exact}

To directly validate the temporal-parallel Jacobi approximation as a training mechanism, we compare it against \emph{exact} sequential recurrent training at three scales. We train \ours models with $\lstart=8$ at 135M, 360M, and 1B scale on the first 1B tokens of FineWeb-Edu. For Jacobi-approximation training we use $(\dforward,\dbackward)=(16,4)$, the same configuration as our main pretraining runs. Exact training with full $2048$-token recurrent backpropagation is computationally intractable, so we use the same truncated-BPTT budget $\dbackward=4$ for a controlled comparison.

As shown in \cref{tab:jacobi_vs_exact}, Jacobi-approximation training closely matches exact recurrent training across all three scales: the Jacobi-trained models are within $0.0010$, $0.0042$, and $0.0045$ validation loss of the exact-recurrence-trained models at 135M, 360M, and 1B respectively (and are in fact marginally better in these runs). Moreover, evaluating a Jacobi-trained model with the approximate forward yields essentially the same perplexity as evaluating it with exact recurrent rollout, consistent with the inference-side measurement in \cref{tab:prefill_overhead}. This indicates that the Jacobi approximation does not introduce meaningful degradation in learned model quality at this scale.

\begin{table}[H]
\centering
\begin{adjustbox}{width=\textwidth}
\begin{tabular}{l|cccc}
\toprule
\toprule
Size & Vanilla Transformer & \ours: exact train, exact eval & \ours: Jacobi train, exact eval & \ours: Jacobi train, Jacobi eval \\
\midrule
135M & 3.9514 (52.01) & 3.9083 (49.81) & \textbf{3.9073} (49.76) & 3.9073 (49.77) \\
360M & 3.3803 (29.38) & 3.3246 (27.79) & \textbf{3.3204} (27.67) & 3.3205 (27.68) \\
1B   & 3.1562 (23.48) & 3.1148 (22.53) & \textbf{3.1103} (22.43) & 3.1102 (22.43) \\
\bottomrule
\end{tabular}
\end{adjustbox}
\caption{Jacobi-approximation training vs.\ exact recurrent training ($\lstart=8$, first 1B FineWeb-Edu tokens, $(\dforward,\dbackward)=(16,4)$). Numbers outside parentheses are validation loss; numbers in parentheses are validation perplexity. Jacobi training matches exact recurrent training across scales, and Jacobi-forward evaluation matches exact recurrent rollout.}
\label{tab:jacobi_vs_exact}
\end{table}

\subsection{Inference Overhead of \ours}
\label{sec:apdx-inference-overhead}

A central claim of \ours is that, unlike looped or pause-token models, it adds negligible per-token inference cost. We verify this empirically. Because the recurrent states are small, all \ours variants add less than $0.1\%$ of the peak GPU memory at inference.

\paragraph{Autoregressive generation.}
We measure free-form generation from the BOS token up to $512$, $1024$, and $2048$ tokens, and report wall-clock time relative to the parameter-matched Transformer baseline (\cref{tab:gen_overhead}). The per-token overhead stays under $\sim$8\% and \emph{decreases} with longer generation and larger models, as the constant gate cost is increasingly dominated by the growing attention cost.

\begin{table}[H]
\centering
\begin{adjustbox}{width=\textwidth}
\begin{tabular}{l|ccccc}
\toprule
\toprule
\# Generated Tokens & 135M \ours(1,30) & 135M \ours(5,26) & 135M \ours(9,22) & 361M \ours(9,24) & 1B \ours(9,24) \\
\midrule
512  & $1.070\times$ & $1.076\times$ & $1.082\times$ & $1.065\times$ & $1.067\times$ \\
1024 & $1.064\times$ & $1.068\times$ & $1.078\times$ & $1.057\times$ & $1.057\times$ \\
2048 & $1.067\times$ & $1.076\times$ & $1.078\times$ & $1.056\times$ & $1.041\times$ \\
\bottomrule
\end{tabular}
\end{adjustbox}
\caption{Autoregressive generation time of \ours relative to the parameter-matched Transformer baseline ($1.0\times$). Overhead is at most $\sim$8\% and decreases with longer generation length and larger model size.}
\label{tab:gen_overhead}
\end{table}

\paragraph{Prefill.}
The main paper reports statistics under exact sequential prompt prefill. The Jacobi approximation can also be applied during prefill, trading a small approximation error for substantially lower prefill latency. \cref{tab:prefill_overhead} reports zero-shot performance together with the prefill speedup over exact prefill at varying Jacobi forward depths $\dforward$. At $\dforward=8$ or $16$, approximate prefill recovers exact-prefill performance to within roughly $0.1\%$ while accelerating prefill by several-fold.

\begin{table}[H]
\centering
\begin{adjustbox}{width=0.85\textwidth}
\begin{tabular}{l|ccccc}
\toprule
\toprule
Model & $\dforward{=}4$ & $\dforward{=}8$ & $\dforward{=}16$ & $\dforward{=}32$ & exact \\
\midrule
\ours(5,26)  & 0.436 ($3.6\times$) & 0.436 ($6.2\times$) & 0.435 ($11.1\times$) & 0.436 ($20.5\times$) & 0.437 \\
\ours(9,22)  & 0.436 ($2.7\times$) & 0.441 ($4.6\times$) & 0.439 ($8.0\times$)  & 0.440 ($14.5\times$) & 0.440 \\
\ours(13,18) & 0.434 ($1.8\times$) & 0.440 ($2.7\times$) & 0.441 ($4.3\times$)  & 0.441 ($7.7\times$)  & 0.441 \\
\ours(15,16) & 0.432 ($1.3\times$) & 0.430 ($1.7\times$) & 0.430 ($2.5\times$)  & 0.429 ($4.0\times$)  & 0.429 \\
Baseline     & --                  & --                  & --                   & --                   & 0.428 \\
\bottomrule
\end{tabular}
\end{adjustbox}
\caption{Approximate (Jacobi) vs.\ exact prefill. Each cell reports zero-shot average performance, with the prefill speedup over exact prefill in parentheses. At $\dforward=8$ or $16$, approximate prefill matches exact-prefill performance within $\sim$0.1\% while being several-fold faster.}
\label{tab:prefill_overhead}
\end{table}

%% file: Appendix/Sections/additional_results.tex
\section{Additional Empirical Results}
\label{sec:apdx-additional-results}

This section reports the additional empirical results discussed in the main text: scaling \ours to larger models and more data, the recurrence-location ablation, and a training-compute-matched baseline.

\subsection{Scaling to Larger Models and More Data}
\label{sec:apdx-scaling}

To test whether the gains of middle-layer recurrence persist beyond the 135M from-scratch setting reported in the main text (\cref{tab:eval_135m}), we pretrain larger \ours variants at 361M and 1B parameters, and additionally extend pretraining up to 50B tokens. We use the gated fusion module with Jacobi depths $(\dforward{=}16, \dbackward{=}4)$ and recurrence starting at $\lstart{=}8$ (denoted gated-f16b4-L8). As summarized in \cref{tab:eval_scale}, the zero-shot NLP gains remain consistent at 361M and 1B scale, and the improvements on reasoning-oriented downstream tasks (HotpotQA-Easy, GSM-Aug) become substantially larger, with $8$--$16\%$ relative gains over the parameter-matched baseline. Extending pretraining to 50B tokens (\cref{tab:eval_50b}) yields even larger gains, especially at 361M scale.

\begin{table}[H]
\centering
\begin{adjustbox}{width=\textwidth}
\begin{tabular}{l|ccccccc|c}
\toprule
\toprule
Model/Config & ARC-C & ARC-E & HS & OBQA & PIQA & SciQ & WG & Average \\
Metric & acc\_n $\uparrow$ & acc\_n $\uparrow$ & acc\_n $\uparrow$ & acc\_n $\uparrow$ & acc\_n $\uparrow$ & acc\_n $\uparrow$ & acc $\uparrow$ & - $\uparrow$ \\
\midrule
\multicolumn{9}{l}{\textit{$\sim$370M parameters}} \\
Baseline (367.8M) & 27.13 & \textbf{52.65} & 37.18 & \textbf{34.00} & 65.51 & 71.00 & 51.85 & 48.48 \\
\ours\ (361.8M; gated-f16b4-L8) & \textbf{28.84} & 52.48 & \textbf{38.66} & 33.40 & \textbf{66.00} & \textbf{73.10} & \textbf{52.09} & \textbf{49.22} \\
\midrule
\multicolumn{9}{l}{\textit{$\sim$1B parameters}} \\
Baseline (996.9M) & 30.97 & \textbf{55.85} & 42.91 & 33.40 & 67.74 & \textbf{77.10} & 51.70 & 51.38 \\
\ours\ (981.5M; gated-f16b4-L8) & \textbf{32.00} & 55.30 & \textbf{45.11} & \textbf{35.40} & \textbf{68.99} & 75.60 & \textbf{54.22} & \textbf{52.37} \\
\bottomrule
\end{tabular}
\end{adjustbox}

\vspace{0.06in}

\begin{adjustbox}{width=\textwidth}
\begin{tabular}{l|ccc}
\toprule
\toprule
Model/Config & HotpotQA-Easy & GSM-Aug-NL & GSM-Aug-Sym \\
Metric & acc $\uparrow$ & acc $\uparrow$ & acc $\uparrow$ \\
\midrule
\multicolumn{4}{l}{\textit{$\sim$370M parameters}} \\
Baseline (367.8M) & 24.43 & 31.08 & 39.35 \\
\ours\ (361.8M; gated-f16b4-L8) & \textbf{28.28 {\scriptsize (+15.8\%)}} & \textbf{34.12 {\scriptsize (+9.8\%)}} & \textbf{42.46 {\scriptsize (+7.9\%)}} \\
\midrule
\multicolumn{4}{l}{\textit{$\sim$1B parameters}} \\
Baseline (996.9M) & 23.28 & 32.37 & 43.97 \\
\ours\ (981.5M; gated-f16b4-L8) & \textbf{26.52 {\scriptsize (+14.0\%)}} & \textbf{36.69 {\scriptsize (+13.3\%)}} & \textbf{44.96 {\scriptsize (+2.3\%)}} \\
\bottomrule
\end{tabular}
\end{adjustbox}
\caption{Scaling \ours to 361M and 1B parameters (pretrained on 10B FineWeb-Edu tokens). \textbf{Top:} zero-shot NLP benchmarks. \textbf{Bottom:} reasoning downstream tasks (HotpotQA-Easy, GSM-Aug natural-language and symbolic), with relative improvement over the baseline in parentheses.}
\label{tab:eval_scale}
\end{table}

\begin{table}[H]
\centering
\begin{adjustbox}{width=\textwidth}
\begin{tabular}{l|ccccccc|c}
\toprule
\toprule
Model/Config & ARC-C & ARC-E & HS & OBQA & PIQA & SciQ & WG & Average \\
Metric & acc\_n $\uparrow$ & acc\_n $\uparrow$ & acc\_n $\uparrow$ & acc\_n $\uparrow$ & acc\_n $\uparrow$ & acc\_n $\uparrow$ & acc $\uparrow$ & - $\uparrow$ \\
\midrule
\multicolumn{9}{l}{\textit{$\sim$135M parameters}} \\
Baseline & \textbf{27.82} & \textbf{52.53} & 36.82 & 33.60 & 65.67 & 72.40 & 49.64 & 48.35 \\
\ours\ (9,22) & 27.56 & 51.30 & \textbf{38.74} & \textbf{34.00} & \textbf{66.10} & \textbf{75.20} & \textbf{53.04} & \textbf{49.42} \\
\midrule
\multicolumn{9}{l}{\textit{$\sim$370M parameters}} \\
Baseline & 31.23 & \textbf{59.64} & 46.28 & 34.60 & \textbf{70.08} & 76.00 & 52.01 & 52.83 \\
\ours\ (9,24) & \textbf{33.28} & 58.88 & \textbf{48.57} & \textbf{36.80} & 69.42 & \textbf{80.30} & \textbf{56.20} & \textbf{54.78} \\
\bottomrule
\end{tabular}
\end{adjustbox}

\vspace{0.06in}

\begin{adjustbox}{width=0.62\textwidth}
\begin{tabular}{l|cc}
\toprule
\toprule
Model/Config & GSM-Aug-NL & GSM-Aug-Sym \\
Metric & acc $\uparrow$ & acc $\uparrow$ \\
\midrule
\multicolumn{3}{l}{\textit{$\sim$135M parameters}} \\
Baseline & 26.46 & 38.67 \\
\ours\ (9,22) & \textbf{30.48 {\scriptsize (+15.2\%)}} & \textbf{41.55 {\scriptsize (+7.4\%)}} \\
\midrule
\multicolumn{3}{l}{\textit{$\sim$370M parameters}} \\
Baseline & 40.18 & 42.53 \\
\ours\ (9,24) & \textbf{44.35 {\scriptsize (+10.4\%)}} & \textbf{46.63 {\scriptsize (+9.6\%)}} \\
\bottomrule
\end{tabular}
\end{adjustbox}
\caption{Extending pretraining to 50B FineWeb-Edu tokens. \textbf{Top:} zero-shot NLP benchmarks for 135M and 361M models. \textbf{Bottom:} grade-school math reasoning downstream (GSM-Aug). Training on more data yields larger gains, especially at 361M scale (average $52.83 \to 54.78$).}
\label{tab:eval_50b}
\end{table}

\paragraph{Recurrence location ablation.}
The scaling configurations above couple recurrence \emph{width} ($D$) with recurrence \emph{location}. To isolate the effect of location, we run a fixed-width ablation that places the recurrent block over early, middle, or late layers while holding $D$ constant. As shown in \cref{tab:layer_location}, at both $D{=}6$ and $D{=}14$, middle-layer recurrence clearly outperforms placing the same recurrent block over the earliest or latest layers. This supports our motivation (\cref{sec:motivation}) that the middle layers, which host the most abstract computation \citep{skean2025layerbylayer,saunshi2024inductive}, are where temporal recurrence is most beneficial.

\begin{table}[H]
\centering
\begin{adjustbox}{width=\textwidth}
\begin{tabular}{l|ccccccc|c}
\toprule
\toprule
Model/Config & ARC-C & ARC-E & HS & OBQA & PIQA & SciQ & WG & Average \\
Metric & acc\_n $\uparrow$ & acc\_n $\uparrow$ & acc\_n $\uparrow$ & acc\_n $\uparrow$ & acc\_n $\uparrow$ & acc\_n $\uparrow$ & acc $\uparrow$ & - $\uparrow$ \\
\midrule
Baseline & 24.74 & 44.28 & 29.81 & 30.20 & 61.53 & 60.80 & 48.46 & 42.83 \\
\midrule
\multicolumn{9}{l}{\textit{6-Layer Recurrence ($D{=}6$)}} \\
13--18 (Middle) & 24.23 & \textbf{45.50} & \textbf{29.95} & \textbf{31.20} & \textbf{61.70} & \textbf{64.20} & \textbf{52.17} & \textbf{44.14} \\
1--6 (Early) & 24.66 & 45.08 & 29.83 & 30.00 & 60.07 & 61.00 & 48.54 & 42.74 \\
25--30 (Late) & \textbf{25.09} & 45.24 & 29.65 & 28.20 & 60.12 & 61.20 & 50.59 & 42.87 \\
\midrule
\multicolumn{9}{l}{\textit{14-Layer Recurrence ($D{=}14$)}} \\
9--22 (Middle) & \textbf{24.15} & \textbf{45.24} & \textbf{30.40} & 29.20 & \textbf{61.15} & \textbf{66.40} & \textbf{51.54} & \textbf{44.01} \\
1--14 (Early) & 23.38 & 44.70 & 28.59 & 29.00 & 59.25 & 59.50 & 51.07 & 42.21 \\
17--30 (Late) & 23.98 & 44.02 & 29.97 & \textbf{31.40} & 60.28 & 61.90 & 50.43 & 43.14 \\
\bottomrule
\end{tabular}
\end{adjustbox}
\caption{Fixed-width recurrence-location ablation at 135M (10B tokens). Holding the recurrence depth $D$ constant, middle-layer recurrence outperforms placing the same recurrent block over the earliest or latest layers, at both $D{=}6$ and $D{=}14$.}
\label{tab:layer_location}
\end{table}

\subsection{Training-Compute-Matched Baseline}
\label{sec:apdx-compute-matched}

Our main comparisons are matched on parameter count, data, and inference compute, but \emph{not} on training wall-clock: the Jacobi-style approximation makes \ours training slower than the baseline (\cref{sec:apdx-compute-overhead}). For a training-compute-matched comparison, we additionally train the Transformer baseline for 2.24 epochs, matching the $2.24\times$ training wall-clock of \ours(13,18). As shown in \cref{tab:compute_matched}, under this matched training budget the longer-trained Transformer surpasses the 135M \ours on zero-shot NLP. This is consistent with our positioning of \ours as an \emph{inference-side and architectural} contribution: under fixed parameter / data / inference-compute budgets it adds a temporal latent pathway with little decoding cost (at most $\sim$8\% per-token overhead) and a more pronounced boost to state-tracking and multi-hop reasoning, at the cost of additional training compute.

\begin{table}[H]
\centering
\begin{adjustbox}{width=\textwidth}
\begin{tabular}{l|ccccccc|c}
\toprule
\toprule
Model/Config & ARC-C & ARC-E & HS & OBQA & PIQA & SciQ & WG & Average \\
Metric & acc\_n $\uparrow$ & acc\_n $\uparrow$ & acc\_n $\uparrow$ & acc\_n $\uparrow$ & acc\_n $\uparrow$ & acc\_n $\uparrow$ & acc $\uparrow$ & - $\uparrow$ \\
\midrule
Transformer (1 epoch) & 24.74 & 44.28 & 29.81 & 30.20 & 61.53 & 60.80 & 48.46 & 42.83 \\
\ours(13,18) & 24.23 & 45.50 & 29.95 & 31.20 & 61.70 & 64.20 & 52.17 & 44.14 \\
Transformer (2.24 epochs) & \textbf{25.68} & \textbf{46.30} & \textbf{32.15} & \textbf{31.80} & \textbf{62.40} & \textbf{66.50} & \textbf{52.25} & \textbf{45.30} \\
\bottomrule
\end{tabular}
\end{adjustbox}
\caption{Training-compute-matched comparison at 135M. The Transformer baseline trained for 2.24 epochs matches the $2.24\times$ training wall-clock of \ours(13,18). Under matched training compute, longer Transformer training surpasses \ours on zero-shot NLP; \ours's advantage is at fixed parameter / data / inference-compute budgets and on reasoning tasks.}
\label{tab:compute_matched}
\end{table}

%% file: Appendix/Sections/future_token_prediction.tex
\section{Future Token Prediction}
\label{sec:mechanistic-analysis}
Temporal middle-layer recurrence creates an explicit temporal shortcut: the recurrent cache computed from token $t$ is fused into the residual stream when processing token $t{+}1$, so the loss at position $t{+}1$ can backpropagate through this pathway into token $t$'s intermediate representation.

This encourages the model to maintain latent states that better anticipate upcoming tokens. To test this hypothesis, we perform a probing analysis: from each intermediate layer, we extract the representation of token $t$ and train a two-layer MLP to predict the next token $x_{t+1}$. Table~\ref{tab:future-token-prediction} reports next-token prediction loss for \ours\ and a parameter-matched Transformer baseline under two choices of $\lend$. For this analysis, negative indices count backward from the LM head, so $\lend=-k$ places the recurrence endpoint $k$ layers below the head. We probe representations from layers $\ell_{-k}$ under the same convention.

\begin{table}[th]
\centering
\begin{tabular}{lcccccccc}
\toprule
& \multicolumn{3}{c}{$\lend = -3$} & \multicolumn{5}{c}{$\lend = -5$} \\
\cmidrule(lr){2-4} \cmidrule(lr){5-9}
 & $\ell_{-1}$ & $\ell_{-2}$ & $\ell_{-3}$ & $\ell_{-1}$ & $\ell_{-2}$ & $\ell_{-3}$ & $\ell_{-4}$ & $\ell_{-5}$ \\
\midrule
\ours  & $\mathbf{5.091}$ & $\mathbf{5.088}$ & $\mathbf{5.110}$ & $\mathbf{5.097}$ & $\mathbf{5.096}$ & $\mathbf{5.109}$ & $\mathbf{5.126}$ & $\mathbf{5.158}$ \\
Baseline & $5.135$ & $5.131$ & $5.145$ & $5.135$ & $5.131$ & $5.145$ & $5.173$ & $5.182$ \\
\bottomrule
\end{tabular}
\caption{Each predictor is trained for 20,000 steps on Fineweb-edu 10B.}
\label{tab:future-token-prediction}
\end{table}

Table~\ref{tab:future-token-prediction} shows that intermediate-layer representations learned by \ours consistently yield lower next-token prediction loss than the parameter-matched baseline, providing direct evidence that \ours promotes representations that better anticipate future tokens.

%% file: Appendix/Sections/training_analysis.tex
\section{Extended Discussions}

\subsection{\ours as a generalized Transformer and recurrent model}

\ours subsumes both recurrent neural networks and the standard Transformer as special cases. In particular, when the learned self-attention pattern reduces to the identity—so that token interactions occur exclusively through recurrence—\ours becomes a fully recurrent model. Conversely, when the learned gating parameters $\gcur$ and $\grec$ are set to zero, the recurrent pathway is disabled and \ours recovers the standard Transformer architecture.

Consequently, \ours has the expressive capacity to smoothly interpolate between recurrent and attention-based computation. Since Transformers excel at in-context retrieval tasks \citep{jelassi2024repeat, nichani2025associative} while recurrent models are well suited for explicit state tracking \citep{merrill2024illusionstatestatespacemodels,grazzi2025unlocking}, \ours has the expressivity to naturally support both modes of computation within a unified framework.

\subsection{Learned gating behavior}


As formulated in Section~\ref{sec:gating}, \ours learns scalar, data-independent gates $\gcur$ and $\grec$ that control the contributions of the input and recurrent streams, respectively. To better understand how the model allocates these streams relative to the residual branch $\vh_t^{(\lstart-1)}$ in Equation~\ref{eq:gating}, we analyze the learned gating parameters over the course of training.

\begin{figure}[h]
    \centering
    \includegraphics[width=0.45\textwidth]{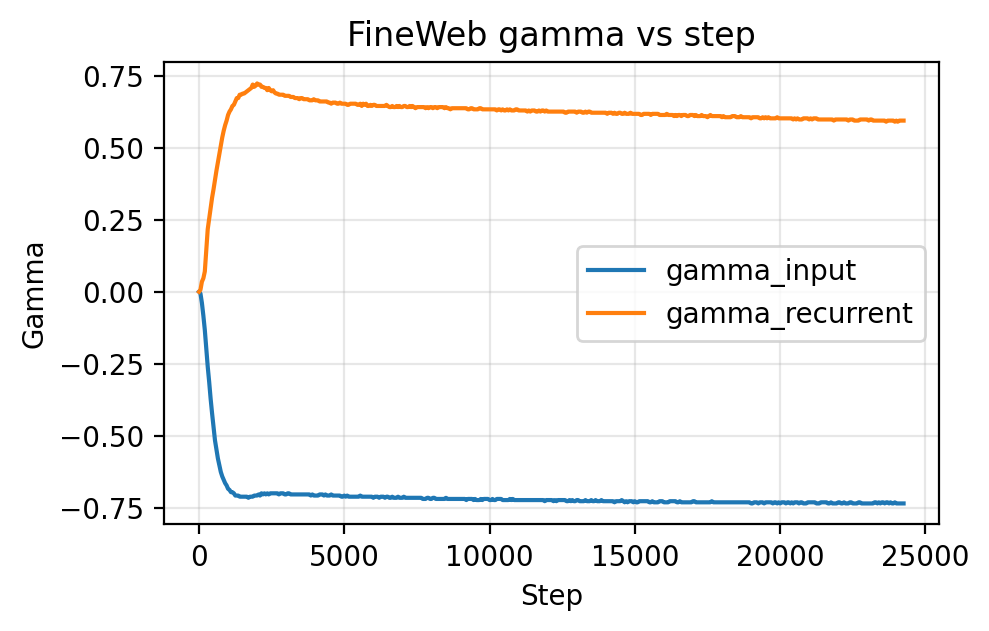}
    \caption{Plot of gating parameters $\gcur$ and $\grec$ with respect to training steps on the FineWeb dataset. This illustrates the learned behavior of input and recurrent gating over the course of training.}
    \label{fig:fineweb-gamma-step}
\end{figure}

Empirically, across all tasks we study, the recurrent gate $\grec$ consistently converges to positive values, while the input gate $\gcur$ consistently converges to negative values. Figure~\ref{fig:fineweb-gamma-step} illustrates this behavior for pretraining on FineWeb. This pattern aligns with the structure of Equation~\ref{eq:gating}: since the input representation $\vh_t^{(\lstart-1)}$ is already added unconditionally via the residual connection, the role of the $\gamma$-gated terms is not to reintroduce the input signal, but rather to \emph{modulate deviations from the identity mapping}. In particular, a negative $\gcur$ suppresses redundant amplification of the input stream, while a positive $\grec$ selectively promotes the recurrent contribution, allowing information from previous time steps to be injected only when it provides additional predictive value beyond the current-token representation.
This asymmetry suggests that \ours learns to treat recurrence as an additive refinement to the residual pathway, rather than as a competing source of information. In effect, the model preserves the standard Transformer computation as a default and leverages the recurrent stream to inject latent, temporally accumulated information when beneficial.

Finally, note that both $\gcur$ and $\grec$ are initialized to zero, so the gating module in Equation~\ref{eq:gating} initially reduces to the identity function. As a result, \ours starts training in a regime that is exactly equivalent to a standard Transformer, and only gradually departs from it as the gating parameters adapt. This design stabilizes optimization and ensures that recurrent computation is introduced smoothly, rather than being imposed a priori.


%% file: Appendix/Sections/experiment_setup.tex
\section{Additional Experiment Setups and Results}

\subsection{NLP benchmarks for evaluating pre-trained models}

We use the following abbreviations for the NLP benchmarks: ARC-C/E: ARC-Challenge/Easy \citep{clark2018think}, HS: HellaSwag \citep{zellers2019hellaswag}, OBQA: OpenBookQA \citep{mihaylov2018can}, PIQA: PhysicalInteractionQA \citep{bisk2020piqa}, SciQ: ScienceExamQA \citep{welbl2017crowdsourcing}, WG: Winogrande \citep{sakaguchi2021winogrande}.

This set of benchmarks follows standard evaluation setup in the pretraining literature.

\subsection{Data Curation for HotpotQA}
To construct SFT data with step-by-step multi-hop reasoning, we curate a subset of HotpotQA and generate 19k chain-of-thought reasoning traces using GPT-4o-mini. As noted earlier, we restrict our experiments to the \textit{easy} subset, since preliminary experiments show that both baseline models and \ours struggle significantly on the medium and hard subsets, making it difficult to draw meaningful comparisons.

\begin{tcolorbox}[title=Prompt for generating HotpotQA chain of thoughts]
According to the context, answer the question in a multi-hop reasoning process. Be sure to include a detailed and clear logic chain in your reasoning. Detail your thought process at each step. Put your final answer within \text{\textbackslash boxed\{\}}. Output example: \text{\textbackslash boxed\{Knox County Regional Airport\}}.

Question: \{\textit{question}\}

Context: \{\textit{context}\}

Note that the correct final answer is \text{\textbackslash boxed\{\textit{answer}\}}.
\end{tcolorbox}

\subsection{Training details}
\label{sec:apdx-training_details}
We note the training details used for each task. 

\begin{table}[H]
\centering
\begin{tabular}{l|l l l l }
\toprule
\textbf{Task} & \textbf{Model} & \textbf{LR} & \textbf{Schedule} & \textbf{Steps} \\
\midrule
S5-Retrieval & 
\begin{tabular}[c]{@{}l@{}}
\textbf{Transformer:} 4-layer, 6-head \\
384 hidden dim, RoPE\\
\textbf{LSTM:} 4-layer,\\
456 hidden size
\end{tabular} & 
5e-4 & Cosine & 200k \\
\midrule
FineWeb-Edu Pretraining    & SmolLM2-135M-\ours          & 5e-4 & Cosine w/ min LR & ~20k  \\\midrule
GSM8K-Aug  & SmolLM2-135M-\ours & 5e-5 & Cosine & 8000 \\
\midrule
HotpotQA & SmolLM2-135M-\ours & 1e-4 & Cosine w/ min LR & 5000 \\
\bottomrule
\end{tabular}
\caption{Training details for various experiments.}
\end{table}

For pretraining on FineWeb-Edu, we used AdamW \citep{loshchilov2018decoupled} optimizer with $\beta_1=0.9$, $\beta_2=0.98$, and weight-decay of $\lambda=0.01$. We set the minimum learning rate decay to be $0.001$ times the peak learning rate. We pack the pretraining data into sequences of length 2048 and train with batchsize 256.
For finetuning runs we used AdamW \citep{loshchilov2018decoupled} optimizer with $\beta_1=0.9$, $\beta_2=0.999$, and weight-decay of $\lambda=0.01$.
Training is conducted on 4 $\times$ H100 GPUs.
\newpage

\paragraph{Data examples}\label{sec:data-examples}
We provide simple examples of the datasets used in our tasks.

\begin{table}[H]
\centering
\renewcommand{\arraystretch}{1.05}
\begin{adjustbox}{width=\textwidth}
\begin{tabular}{p{2.8cm}|p{11.2cm}}
\toprule
\textbf{Dataset} & \textbf{Details} \\
\midrule
ProsQA-Hard & {
\textbf{Input:} Every zolufibus is a conarus. Every conarus is a tanirus. Every tanirus is a gedatus.
Every cemapus is a lovaperus. Every lovaperus is a zucirarus. Every zucirarus is a pilevus.
Every sasarus is a madusurus. Every madusurus is a cidodivus. Every cidodivus is a bumobus.
Every bumobus is a nuronus. Every nuronus is a metuvelus. Every metuvelus is a resamus.
Every resamus is a gedatus.
Jordan is a sasarus.
Is Jordan a gedatus or a pilevus?

\vspace{0.05in}
\textbf{Steps:} Jordan is a sasarus.\newline
Every sasarus is a madusurus.\newline
Every madusurus is a cidodivus.\newline
Every cidodivus is a bumobus.\newline
Every bumobus is a nuronus.\newline
Every nuronus is a metuvelus.\newline
Every metuvelus is a resamus.\newline
Every resamus is a gedatus.

\vspace{0.05in}
\textbf{Output:} Jordan is a gedatus.
}\\
\midrule
S5 Retrieval & {
\textbf{Input:}\newline
\texttt{<A\_32514>F4Nd <A\_12543>9O8W <A\_54213>Jccq}\newline
\texttt{| <A\_54213> <A\_43125> <A\_52314>}

\vspace{0.05in}
\textbf{Target:}\newline
\texttt{<A\_32514>F4Nd <A\_12543>9O8W <A\_54213>Jccq}\newline
\texttt{| <A\_54213>Jccq <A\_12543>9O8W <A\_32514>F4Nd}
}\\
\midrule
GSM-Aug Natural Language & {
\textbf{Question:} Out of 600 employees in a company, 30\% got promoted while 10\% received bonus. How many employees did not get either a promotion or a bonus?

\vspace{0.05in}
\textbf{Steps:}
600 x 30/100 = 180 employees were promoted.\newline
600 x 10/100 = 60 employees received a bonus.\newline
So a total of 180+60=240 employees received a promotion or a bonus.\newline
Therefore, 600 - 240 = 360 employees did not get either a promotion or a bonus.

\vspace{0.05in}
\textbf{Answer:} 360 
}\\
\midrule
Variable Assignment & {
\textbf{Input:} Variable assignment. Follow the assignments and fill in the blank.\newline
a=5\newline
b=2\newline
c=8\newline
x=c\newline
y=x\newline
z=y\newline
w=b\newline
v=w\newline
z=\_\_\_\newline
Answer: |

\vspace{0.05in}
\textbf{Outputs:}
8
}\\
\midrule
HotpotQA & {
\textbf{Input:}
Which magazine was founded first, \emph{The Economist} or \emph{The Atlantic}?

\vspace{0.05in}
\textbf{Context:}\newline
Document 1: \emph{The Economist} is a weekly newspaper founded in 1843.\newline
Document 2: \emph{The Atlantic} is a magazine founded in 1857.

\vspace{0.05in}
\textbf{Steps:}\newline
The Economist was founded in 1843.\newline
The Atlantic was founded in 1857.\newline
1843 is earlier than 1857.

\vspace{0.05in}
\textbf{Answer:} The Economist
}\\
\bottomrule
\end{tabular}
\end{adjustbox}

\caption{Examples for each dataset used.}
\label{tab:task_example}
\end{table}
\newpage

\paragraph{Additional Pretraining Results}
Table~\ref{tab:perplexity} shows the pretraining perplexities of \ours and baseline Transformer at different model scales.

\begin{table}[th]
\centering
\begin{tabular}{lccccc}
\toprule
 & \multicolumn{2}{c}{135M} & \multicolumn{2}{c}{368M} \\
\cmidrule(lr){2-3} \cmidrule(lr){4-5}
  & Baseline & \ours & Baseline & \ours \\
\midrule
Perplexity$\downarrow$ & 22.64 & \textbf{21.54} & 16.12 & \textbf{15.79} \\
\bottomrule
\end{tabular}
\caption{We pretrain the standard Transformer and \ours, both based on SmolLM, at different scales on Fineweb-edu-10B, with \ours consistently having improved test time performance.}
\label{tab:perplexity}
\end{table}


%% file: colm2026_conference.bib
@misc{yue2025hybridlatentreasoningreinforcement,
      title={Hybrid Latent Reasoning via Reinforcement Learning}, 
      author={Zhenrui Yue and Bowen Jin and Huimin Zeng and Honglei Zhuang and Zhen Qin and Jinsung Yoon and Lanyu Shang and Jiawei Han and Dong Wang},
      year={2025},
      eprint={2505.18454},
      archivePrefix={arXiv},
      primaryClass={cs.CL},
      url={https://arxiv.org/abs/2505.18454}, 
}

@inproceedings{tenney2019bert,
  title={BERT Rediscovers the Classical NLP Pipeline},
  author={Tenney, Ian and Das, Dipanjan and Pavlick, Ellie},
  booktitle={ACL},
  year={2019}
}

@article{geva2021transformer,
  title={Transformer Feed-Forward Layers Are Key-Value Memories},
  author={Geva, Mor and Schuster, Roei and Berant, Jonathan and Levy, Omer},
  journal={EMNLP},
  year={2021}
}

@article{meng2022locating,
  title={Locating and Editing Factual Associations in GPT},
  author={Meng, Kevin and Bau, David and Andonian, Alex and Belinkov, Yonatan},
  journal={NeurIPS},
  year={2022}
}

@misc{yang2024latently,
  title         = {Do Large Language Models Latently Perform Multi-Hop Reasoning?},
  author        = {Yang, Sohee and Gribovskaya, Elena and Kassner, Nora and Geva, Mor and Riedel, Sebastian},
  year          = {2024},
  eprint        = {2402.16837},
  archivePrefix = {arXiv},
  primaryClass  = {cs.CL},
  url           = {https://arxiv.org/abs/2402.16837}
}

@misc{deng2023implicitcot,
  title         = {Implicit Chain of Thought Reasoning via Knowledge Distillation},
  author        = {Deng, Yuntian and Prasad, Kiran and Fernandez, Roland and Smolensky, Paul and Chaudhary, Vishrav and Shieber, Stuart},
  year          = {2023},
  eprint        = {2311.01460},
  archivePrefix = {arXiv},
  primaryClass  = {cs.CL},
  url           = {https://arxiv.org/abs/2311.01460}
}

@misc{deng2024explicit2implicit,
  title         = {From Explicit CoT to Implicit CoT: Learning to Internalize CoT Step by Step},
  author        = {Deng, Yuntian and Choi, Yejin and Shieber, Stuart},
  year          = {2024},
  eprint        = {2405.14838},
  archivePrefix = {arXiv},
  primaryClass  = {cs.CL},
  url           = {https://arxiv.org/abs/2405.14838}
}

@misc{merchant2023pause,
  title         = {Think before you speak: Training Language Models With Pause Tokens},
  author        = {Goyal, Sachin and Ji, Ziwei and Rawat, Ankit Singh and Menon, Aditya Krishna and Kumar, Sanjiv and Nagarajan, Vaishnavh},
  year          = {2024},
  eprint        = {2310.02226},
  archivePrefix = {arXiv},
  primaryClass  = {cs.CL},
  url           = {https://arxiv.org/abs/2310.02226}
}

@inproceedings{skean2025layerbylayer,
  title         = {Layer by Layer: Uncovering Hidden Representations in Language Models},
  author        = {Skean, Oscar and Arefin, Md Rifat and Zhao, Dan and Patel, Niket and Naghiyev, Jalal and LeCun, Yann and Shwartz-Ziv, Ravid},
  booktitle     = {Forty-second International Conference on Machine Learning (ICML)},
  year          = {2025},
  url           = {https://openreview.net/forum?id=WGXb7UdvTX}
}

@misc{openmathreasoning2024,
  title         = {AIMO-2 Winning Solution: Building State-of-the-Art Mathematical Reasoning Models with OpenMathReasoning dataset},
  author        = {Moshkov, Ivan and Hanley, Darragh and Sorokin, Ivan and Toshniwal, Shubham and Henkel, Christof and Schifferer, Benedikt and Du, Wei and Gitman, Igor},
  year          = {2025},
  eprint        = {2504.16891},
  archivePrefix = {arXiv},
  primaryClass  = {cs.CL},
  url           = {https://huggingface.co/datasets/nvidia/OpenMathReasoning}
}

@misc{hao2024coconut,
  title         = {Training Large Language Models to Reason in a Continuous Latent Space},
  author        = {Hao, Shibo and Sukhbaatar, Sainbayar and Su, DiJia and Li, Xian and Hu, Zhiting and Weston, Jason and Tian, Yuandong},
  year          = {2024},
  eprint        = {2412.06769},
  archivePrefix = {arXiv},
  primaryClass  = {cs.CL},
  url           = {https://arxiv.org/abs/2412.06769}
}

@misc{tack2025cocomix,
  title         = {LLM Pretraining with Continuous Concepts},
  author        = {Tack, Jihoon and Lanchantin, Jack and Yu, Jane and Cohen, Andrew and Kulikov, Ilia and Lan, Janice and Hao, Shibo and Tian, Yuandong and Weston, Jason and Li, Xian},
  year          = {2025},
  eprint        = {2502.08524},
  archivePrefix = {arXiv},
  primaryClass  = {cs.CL},
  url           = {https://arxiv.org/abs/2502.08524}
}

@misc{shen2025codi,
  title         = {CODI: Compressing Chain-of-Thought into Continuous Space via Self-Distillation},
  author        = {Shen, Zhenyi and Yan, Hanqi and Zhang, Linhai and Hu, Zhanghao and Du, Yali and He, Yulan},
  year          = {2025},
  eprint        = {2502.21074},
  archivePrefix = {arXiv},
  primaryClass  = {cs.CL},
  url           = {https://arxiv.org/abs/2502.21074}
}

@misc{su2025tokenassorted,
  title         = {Token Assorted: Mixing Latent and Text Tokens for Improved Language Model Reasoning},
  author        = {Su, DiJia and Zhu, Hanlin and Xu, Yingchen and Jiao, Jiantao and Tian, Yuandong and Zheng, Qinqing},
  year          = {2025},
  eprint        = {2502.03275},
  archivePrefix = {arXiv},
  primaryClass  = {cs.CL},
  url           = {https://arxiv.org/abs/2502.03275}
}

@misc{kong2025ltm,
  title         = {Latent Thought Models with Variational Bayes Inference-Time Computation},
  author        = {Kong, Deqian and Zhao, Minglu and Xu, Dehong and Pang, Bo and Wang, Shu and Honig, Edouardo and Si, Zhangzhang and Li, Chuan and Xie, Jianwen and Xie, Sirui and Wu, Ying Nian},
  year          = {2025},
  eprint        = {2502.01567},
  archivePrefix = {arXiv},
  primaryClass  = {cs.CL},
  url           = {https://arxiv.org/abs/2502.01567}
}

@misc{dai2019transformerxl,
  title         = {Transformer-XL: Attentive Language Models Beyond a Fixed-Length Context},
  author        = {Dai, Zihang and Yang, Zhilin and Yang, Yiming and Carbonell, Jaime and Le, Quoc V. and Salakhutdinov, Ruslan},
  year          = {2019},
  eprint        = {1901.02860},
  archivePrefix = {arXiv},
  url           = {https://arxiv.org/abs/1901.02860}
}

@misc{rae2019compressive,
  title         = {Compressive Transformers for Long-Range Sequence Modelling},
  author        = {Rae, Jack W. and Potapenko, Anna and Jayakumar, Siddhant M. and Lillicrap, Timothy P.},
  year          = {2019},
  eprint        = {1911.05507},
  archivePrefix = {arXiv},
  url           = {https://arxiv.org/abs/1911.05507}
}

@misc{bulatov2022rmt,
  title         = {Recurrent Memory Transformer},
  author        = {Bulatov, Aydar and Kuratov, Yuri and Burtsev, Mikhail S.},
  year          = {2022},
  eprint        = {2207.06881},
  archivePrefix = {arXiv},
  url           = {https://arxiv.org/abs/2207.06881}
}

@inproceedings{vaswani2017attention,
  title={Attention Is All You Need},
  author={Vaswani, Ashish and Shazeer, Noam and Parmar, Niki and others},
  booktitle={Advances in Neural Information Processing Systems},
  year={2017}
}

@inproceedings{dehghani2018universal,
  title={Universal Transformers},
  author={Dehghani, Mostafa and Gouws, Stephan and others},
  booktitle={International Conference on Learning Representations},
  year={2019}
}

@article{openai2023gpt4,
  title={GPT-4 Technical Report},
  author={{OpenAI}},
  journal={arXiv preprint arXiv:2303.08774},
  year={2023}
}

@article{wei2022chainofthought,
  title={Chain-of-Thought Prompting Elicits Reasoning in Large Language Models},
  author={Wei, Jason and Wang, Xuezhi and Schuurmans, Dale and Bosma, Maarten and Chi, Ed and Le, Quoc and Zhou, Denny},
  journal={arXiv preprint arXiv:2201.11903},
  year={2022}
}

@inproceedings{grazzi2025unlocking,
  title={Unlocking State-Tracking in Linear RNNs Through Negative Eigenvalues},
  author={Riccardo Grazzi and Julien Siems and Arber Zela and Jörg K. H. Franke and Frank Hutter and Massimiliano Pontil},
  booktitle={International Conference on Learning Representations (ICLR)},
  year={2025},
  url={https://openreview.net/forum?id=UvTo3tVBk2}
}

@inproceedings{jelassi2024repeat,
  title={Repeat After Me: Transformers are Better than State Space Models at Copying},
  author={Jelassi, Samy and Brandfonbrener, David and Girotti, Sham and Li, Yuanzhi and Lazaric, Alessandro},
  booktitle={Proceedings of the 41st International Conference on Machine Learning (ICML)},
  year={2024},
  url={https://arxiv.org/abs/2402.01032}
}

@inproceedings{nichani2025associative,
  title={Understanding Factual Recall in Transformers via Associative Memories},
  author={Nichani, Preetum and Vitushinsky, Alex and Mei, Song and Gonen, Hila and Li, Yuanzhi},
  booktitle={International Conference on Learning Representations (ICLR)},
  year={2025},
  note={Spotlight paper},
  url={https://arxiv.org/abs/2406.06484}
}

@misc{behrouz2024titans,
  title         = {Titans: Learning to Memorize at Test Time},
  author        = {Behrouz, Ali and Zhong, Peilin and Mirrokni, Vahab},
  year          = {2024},
  eprint        = {2501.00663},
  archivePrefix = {arXiv},
  url           = {https://arxiv.org/abs/2501.00663}
}

@misc{gu2021s4,
  title         = {Efficiently Modeling Long Sequences with Structured State Spaces},
  author        = {Gu, Albert and Goel, Karan and R{\'e}, Christopher},
  year          = {2021},
  eprint        = {2111.00396},
  archivePrefix = {arXiv},
  url           = {https://arxiv.org/abs/2111.00396}
}

@misc{poli2023hyena,
  title         = {Hyena Hierarchy: Towards Larger Convolutional Language Models},
  author        = {Poli, Michael and Massaroli, Stefano and Nguyen, Eric and Fu, Daniel Y. and Dao, Tri and Baccus, Stephen and Bengio, Yoshua and Ermon, Stefano and R{\'e}, Christopher},
  year          = {2023},
  eprint        = {2302.10866},
  archivePrefix = {arXiv},
  url           = {https://arxiv.org/abs/2302.10866}
}

@misc{gu2023mamba,
  title         = {Mamba: Linear-Time Sequence Modeling with Selective State Spaces},
  author        = {Gu, Albert and Dao, Tri},
  year          = {2023},
  eprint        = {2312.00752},
  archivePrefix = {arXiv},
  url           = {https://arxiv.org/abs/2312.00752}
}

@misc{sun2023retnet,
  title         = {Retentive Network: A Successor to Transformer for Large Language Models},
  author        = {Sun, Yutao and Dong, Li and Huang, Shaohan and Ma, Shuming and Xia, Yuqing and Xue, Jilong and Wang, Jianyong and Wei, Furu},
  year          = {2023},
  eprint        = {2307.08621},
  archivePrefix = {arXiv},
  url           = {https://arxiv.org/abs/2307.08621}
}

@misc{de2024griffin,
  title         = {Griffin: Mixing Gated Linear Recurrences with Local Attention for Efficient Language Models},
  author        = {De, Soham and Smith, Samuel L. and Fernando, Anushan and Botev, Aleksandar and Cristian-Muraru, George and Gu, Albert and Haroun, Ruba and Berrada, Leonard and Chen, Yutian and Srinivasan, Srivatsan and Desjardins, Guillaume and Doucet, Arnaud and Budden, David and Teh, Yee Whye and Pascanu, Razvan and De Freitas, Nando and Gulcehre, Caglar},
  year          = {2024},
  eprint        = {2402.19427},
  archivePrefix = {arXiv},
  url           = {https://arxiv.org/abs/2402.19427}
}

@article{cobbe2021gsm8k,
  title={Training Verifiers to Solve Math Word Problems},
  author={Cobbe, Karl and Kosaraju, Vineet and Bavarian, Mohammad and Chen, Mark and Jun, Heewoo and Kaiser, Lukasz and Plappert, Matthias and Tworek, Jerry and Hilton, Jacob and Nakano, Reiichiro and Hesse, Christopher and Schulman, John},
  journal={arXiv preprint arXiv:2110.14168},
  year={2021}
}

@inproceedings{
merrill2024illusionstatestatespacemodels,
title={The Illusion of State in State-Space Models},
author={William Merrill and Jackson Petty and Ashish Sabharwal},
booktitle={Forty-first International Conference on Machine Learning},
year={2024},
url={https://openreview.net/forum?id=QZgo9JZpLq}
}

@misc{penedo2024finewebdatasetsdecantingweb,
      title={The FineWeb Datasets: Decanting the Web for the Finest Text Data at Scale}, 
      author={Guilherme Penedo and Hynek Kydlíček and Loubna Ben allal and Anton Lozhkov and Margaret Mitchell and Colin Raffel and Leandro Von Werra and Thomas Wolf},
      year={2024},
      eprint={2406.17557},
      archivePrefix={arXiv},
      primaryClass={cs.CL},
      url={https://arxiv.org/abs/2406.17557}, 
}

@misc{allal2025smollm2smolgoesbig,
      title={SmolLM2: When Smol Goes Big -- Data-Centric Training of a Small Language Model}, 
      author={Loubna Ben Allal and Anton Lozhkov and Elie Bakouch and Gabriel Martín Blázquez and Guilherme Penedo and Lewis Tunstall and Andrés Marafioti and Hynek Kydlíček and Agustín Piqueres Lajarín and Vaibhav Srivastav and Joshua Lochner and Caleb Fahlgren and Xuan-Son Nguyen and Clémentine Fourrier and Ben Burtenshaw and Hugo Larcher and Haojun Zhao and Cyril Zakka and Mathieu Morlon and Colin Raffel and Leandro von Werra and Thomas Wolf},
      year={2025},
      eprint={2502.02737},
      archivePrefix={arXiv},
      primaryClass={cs.CL},
      url={https://arxiv.org/abs/2502.02737}, 
}

@inproceedings{
zhang2025soft,
title={Soft Thinking: Unlocking the Reasoning Potential of {LLM}s in Continuous Concept Space},
author={Zhen Zhang and Xuehai He and Weixiang Yan and Ao Shen and Chenyang Zhao and Xin Eric Wang},
booktitle={The Thirty-ninth Annual Conference on Neural Information Processing Systems},
year={2025},
url={https://openreview.net/forum?id=ByQdHPGKgU}
}

@misc{tang2026multiplexthinkingreasoningtokenwise,
      title={Multiplex Thinking: Reasoning via Token-wise Branch-and-Merge}, 
      author={Yao Tang and Li Dong and Yaru Hao and Qingxiu Dong and Furu Wei and Jiatao Gu},
      year={2026},
      eprint={2601.08808},
      archivePrefix={arXiv},
      primaryClass={cs.CL},
      url={https://arxiv.org/abs/2601.08808}, 
}

@misc{butt2025softtokenshardtruths,
      title={Soft Tokens, Hard Truths}, 
      author={Natasha Butt and Ariel Kwiatkowski and Ismail Labiad and Julia Kempe and Yann Ollivier},
      year={2025},
      eprint={2509.19170},
      archivePrefix={arXiv},
      primaryClass={cs.CL},
      url={https://arxiv.org/abs/2509.19170}, 
}

@misc{zou2025latentcollaborationmultiagentsystems,
      title={Latent Collaboration in Multi-Agent Systems}, 
      author={Jiaru Zou and Xiyuan Yang and Ruizhong Qiu and Gaotang Li and Katherine Tieu and Pan Lu and Ke Shen and Hanghang Tong and Yejin Choi and Jingrui He and James Zou and Mengdi Wang and Ling Yang},
      year={2025},
      eprint={2511.20639},
      archivePrefix={arXiv},
      primaryClass={cs.CL},
      url={https://arxiv.org/abs/2511.20639}, 
}

@article{fu2025thinkathard,
  title={Think-at-Hard: Selective Latent Iterations to Improve Reasoning Language Models},
  author={Fu, Tianyu and You, Yichen and Chen, Zekai and Dai, Guohao and Yang, Huazhong and Wang, Yu},
  journal={arXiv preprint arXiv:2511.08577},
  year={2025}
}

@inproceedings{
zhuang2025mixture,
title={Mixture of Inputs: Text Generation Beyond Discrete Token Sampling},
author={Yufan Zhuang and Liyuan Liu and Chandan Singh and Jingbo Shang and Jianfeng Gao},
booktitle={The Thirty-ninth Annual Conference on Neural Information Processing Systems},
year={2025},
url={https://openreview.net/forum?id=l6C6Pw30Gl}
}

@inproceedings{
geiping2025recurrentdepth,
title={Scaling up Test-Time Compute with Latent Reasoning: A Recurrent Depth Approach},
author={Jonas Geiping and Sean Michael McLeish and Neel Jain and John Kirchenbauer and Siddharth Singh and Brian R. Bartoldson and Bhavya Kailkhura and Abhinav Bhatele and Tom Goldstein},
booktitle={The Thirty-ninth Annual Conference on Neural Information Processing Systems},
year={2025},
url={https://openreview.net/forum?id=S3GhJooWIC}
}

@inproceedings{ott-etal-2019-fairseq,
    title = "fairseq: A Fast, Extensible Toolkit for Sequence Modeling",
    author = "Ott, Myle  and
      Edunov, Sergey  and
      Baevski, Alexei  and
      Fan, Angela  and
      Gross, Sam  and
      Ng, Nathan  and
      Grangier, David  and
      Auli, Michael",
    editor = "Ammar, Waleed  and
      Louis, Annie  and
      Mostafazadeh, Nasrin",
    booktitle = "Proceedings of the 2019 Conference of the North {A}merican Chapter of the Association for Computational Linguistics (Demonstrations)",
    month = jun,
    year = "2019",
    address = "Minneapolis, Minnesota",
    publisher = "Association for Computational Linguistics",
    url = "https://aclanthology.org/N19-4009/",
    doi = "10.18653/v1/N19-4009",
    pages = "48--53"
}

@misc{atanas2025modulardatasetdemonstratellm,
      title={A Modular Dataset to Demonstrate LLM Abstraction Capability}, 
      author={Adam Atanas and Kai Liu},
      year={2025},
      eprint={2503.17645},
      archivePrefix={arXiv},
      primaryClass={cs.AI},
      url={https://arxiv.org/abs/2503.17645}, 
}

@inproceedings{lan2019albert,
  title        = {{ALBERT}: A Lite {BERT} for Self-supervised Learning of Language Representations},
  author       = {Lan, Zhenzhong and Chen, Mingda and Goodman, Sebastian and Gimpel, Kevin and Sharma, Piyush and Soricut, Radu},
  booktitle    = {International Conference on Learning Representations (ICLR)},
  year         = {2020},
  note         = {arXiv:1909.11942},
  url          = {https://arxiv.org/abs/1909.11942}
}

@inproceedings{giannou2023looped,
  title        = {Looped Transformers as Programmable Computers},
  author       = {Giannou, Angeliki and Rajput, Shashank and Sohn, Jy-Yong and Lee, Kangwook and Lee, Jason D. and Papailiopoulos, Dimitris},
  booktitle    = {Proceedings of the 40th International Conference on Machine Learning (ICML)},
  year         = {2023},
  series       = {Proceedings of Machine Learning Research},
  volume       = {202},
  pages        = {11398--11442},
  publisher    = {PMLR},
  note         = {arXiv:2301.13196},
  url          = {https://arxiv.org/abs/2301.13196}
}

@inproceedings{saunshi2025latent,
  title        = {Reasoning with Latent Thoughts: On the Power of Looped Transformers},
  author       = {Saunshi, Nikunj and Dikkala, Nishanth and Li, Zhiyuan and Kumar, Sanjiv and Reddi, Sashank J.},
  booktitle    = {International Conference on Learning Representations (ICLR)},
  year         = {2025},
  note         = {arXiv:2502.17416},
  url          = {https://arxiv.org/abs/2502.17416}
}

@article{zhu2025ouro,
  title        = {Scaling Latent Reasoning via Looped Language Models},
  author       = {Zhu, Rui-Jie and Wang, Zixuan and Hua, Kai and Zhang, Tianyu and Li, Ziniu and Que, Haoran and Wei, Boyi and Wen, Zixin and Yin, Fan and Xing, He and Li, Lu and Shi, Jiajun and Ma, Kaijing and Li, Shanda and Kergan, Taylor and Smith, Andrew and Qu, Xingwei and Hui, Mude and Wu, Bohong and Min, Qiyang and Huang, Hongzhi and Zhou, Xun and Ye, Wei and Liu, Jiaheng and Yang, Jian and Shi, Yunfeng and Lin, Chenghua and Zhao, Enduo and Cai, Tianle and Zhang, Ge and Huang, Wenhao and Bengio, Yoshua and Eshraghian, Jason},
  journal      = {arXiv preprint},
  year         = {2025},
  note         = {arXiv:2510.25741},
  url          = {https://arxiv.org/abs/2510.25741}
}

@article{clark2018think,
  title={Think you have solved question answering? try arc, the ai2 reasoning challenge},
  author={Clark, Peter and Cowhey, Isaac and Etzioni, Oren and Khot, Tushar and Sabharwal, Ashish and Schoenick, Carissa and Tafjord, Oyvind},
  journal={arXiv preprint arXiv:1803.05457},
  year={2018}
}

@article{zellers2019hellaswag,
  title={Hellaswag: Can a machine really finish your sentence?},
  author={Zellers, Rowan and Holtzman, Ari and Bisk, Yonatan and Farhadi, Ali and Choi, Yejin},
  journal={arXiv preprint arXiv:1905.07830},
  year={2019}
}

@article{mihaylov2018can,
  title={Can a suit of armor conduct electricity? a new dataset for open book question answering},
  author={Mihaylov, Todor and Clark, Peter and Khot, Tushar and Sabharwal, Ashish},
  journal={arXiv preprint arXiv:1809.02789},
  year={2018}
}

@inproceedings{bisk2020piqa,
  title={Piqa: Reasoning about physical commonsense in natural language},
  author={Bisk, Yonatan and Zellers, Rowan and Gao, Jianfeng and Choi, Yejin and others},
  booktitle={Proceedings of the AAAI conference on artificial intelligence},
  volume={34},
  number={05},
  pages={7432--7439},
  year={2020}
}

@article{welbl2017crowdsourcing,
  title={Crowdsourcing multiple choice science questions},
  author={Welbl, Johannes and Liu, Nelson F and Gardner, Matt},
  journal={arXiv preprint arXiv:1707.06209},
  year={2017}
}

@article{sakaguchi2021winogrande,
  title={Winogrande: An adversarial winograd schema challenge at scale},
  author={Sakaguchi, Keisuke and Bras, Ronan Le and Bhagavatula, Chandra and Choi, Yejin},
  journal={Communications of the ACM},
  volume={64},
  number={9},
  pages={99--106},
  year={2021},
  publisher={ACM New York, NY, USA}
}

@article{saunshi2024inductive,
  title={On the inductive bias of stacking towards improving reasoning},
  author={Saunshi, Nikunj and Karp, Stefani and Krishnan, Shankar and Miryoosefi, Sobhan and Jakkam Reddi, Sashank and Kumar, Sanjiv},
  journal={Advances in Neural Information Processing Systems},
  volume={37},
  pages={71437--71464},
  year={2024}
}

@inproceedings{
liu2023transformers,
title={Transformers Learn Shortcuts to Automata},
author={Bingbin Liu and Jordan T. Ash and Surbhi Goel and Akshay Krishnamurthy and Cyril Zhang},
booktitle={The Eleventh International Conference on Learning Representations },
year={2023},
url={https://openreview.net/forum?id=De4FYqjFueZ}
}

@inproceedings{
li2025how,
title={(How) Do Language Models Track State?},
author={Belinda Z. Li and Zifan Carl Guo and Jacob Andreas},
booktitle={Forty-second International Conference on Machine Learning},
year={2025},
url={https://openreview.net/forum?id=8SXosAVIFH}
}

@article{touvron2023llama,
  title={Llama: Open and efficient foundation language models},
  author={Touvron, Hugo and Lavril, Thibaut and Izacard, Gautier and Martinet, Xavier and Lachaux, Marie-Anne and Lacroix, Timoth{\'e}e and Rozi{\`e}re, Baptiste and Goyal, Naman and Hambro, Eric and Azhar, Faisal and others},
  journal={arXiv preprint arXiv:2302.13971},
  year={2023}
}

@misc{eval-harness,
  author       = {Gao, Leo and Tow, Jonathan and Abbasi, Baber and Biderman, Stella and Black, Sid and DiPofi, Anthony and Foster, Charles and Golding, Laurence and Hsu, Jeffrey and Le Noac'h, Alain and Li, Haonan and McDonell, Kyle and Muennighoff, Niklas and Ociepa, Chris and Phang, Jason and Reynolds, Laria and Schoelkopf, Hailey and Skowron, Aviya and Sutawika, Lintang and Tang, Eric and Thite, Anish and Wang, Ben and Wang, Kevin and Zou, Andy},
  title        = {The Language Model Evaluation Harness},
  month        = 07,
  year         = 2024,
  publisher    = {Zenodo},
  version      = {v0.4.3},
  doi          = {10.5281/zenodo.12608602},
  url          = {https://zenodo.org/records/12608602}
}

@misc{openai2024gpt4ocard,
      title={GPT-4o System Card}, 
      author={OpenAI and : and Aaron Hurst and Adam Lerer and Adam P. Goucher and Adam Perelman and Aditya Ramesh and Aidan Clark and AJ Ostrow and Akila Welihinda and Alan Hayes and Alec Radford and Aleksander Mądry and Alex Baker-Whitcomb and Alex Beutel and Alex Borzunov and Alex Carney and Alex Chow and Alex Kirillov and Alex Nichol and Alex Paino and Alex Renzin and Alex Tachard Passos and Alexander Kirillov and Alexi Christakis and Alexis Conneau and Ali Kamali and Allan Jabri and Allison Moyer and Allison Tam and Amadou Crookes and Amin Tootoochian and Amin Tootoonchian and Ananya Kumar and Andrea Vallone and Andrej Karpathy and Andrew Braunstein and Andrew Cann and Andrew Codispoti and Andrew Galu and Andrew Kondrich and Andrew Tulloch and Andrey Mishchenko and Angela Baek and Angela Jiang and Antoine Pelisse and Antonia Woodford and Anuj Gosalia and Arka Dhar and Ashley Pantuliano and Avi Nayak and Avital Oliver and Barret Zoph and Behrooz Ghorbani and Ben Leimberger and Ben Rossen and Ben Sokolowsky and Ben Wang and Benjamin Zweig and Beth Hoover and Blake Samic and Bob McGrew and Bobby Spero and Bogo Giertler and Bowen Cheng and Brad Lightcap and Brandon Walkin and Brendan Quinn and Brian Guarraci and Brian Hsu and Bright Kellogg and Brydon Eastman and Camillo Lugaresi and Carroll Wainwright and Cary Bassin and Cary Hudson and Casey Chu and Chad Nelson and Chak Li and Chan Jun Shern and Channing Conger and Charlotte Barette and Chelsea Voss and Chen Ding and Cheng Lu and Chong Zhang and Chris Beaumont and Chris Hallacy and Chris Koch and Christian Gibson and Christina Kim and Christine Choi and Christine McLeavey and Christopher Hesse and Claudia Fischer and Clemens Winter and Coley Czarnecki and Colin Jarvis and Colin Wei and Constantin Koumouzelis and Dane Sherburn and Daniel Kappler and Daniel Levin and Daniel Levy and David Carr and David Farhi and David Mely and David Robinson and David Sasaki and Denny Jin and Dev Valladares and Dimitris Tsipras and Doug Li and Duc Phong Nguyen and Duncan Findlay and Edede Oiwoh and Edmund Wong and Ehsan Asdar and Elizabeth Proehl and Elizabeth Yang and Eric Antonow and Eric Kramer and Eric Peterson and Eric Sigler and Eric Wallace and Eugene Brevdo and Evan Mays and Farzad Khorasani and Felipe Petroski Such and Filippo Raso and Francis Zhang and Fred von Lohmann and Freddie Sulit and Gabriel Goh and Gene Oden and Geoff Salmon and Giulio Starace and Greg Brockman and Hadi Salman and Haiming Bao and Haitang Hu and Hannah Wong and Haoyu Wang and Heather Schmidt and Heather Whitney and Heewoo Jun and Hendrik Kirchner and Henrique Ponde de Oliveira Pinto and Hongyu Ren and Huiwen Chang and Hyung Won Chung and Ian Kivlichan and Ian O'Connell and Ian O'Connell and Ian Osband and Ian Silber and Ian Sohl and Ibrahim Okuyucu and Ikai Lan and Ilya Kostrikov and Ilya Sutskever and Ingmar Kanitscheider and Ishaan Gulrajani and Jacob Coxon and Jacob Menick and Jakub Pachocki and James Aung and James Betker and James Crooks and James Lennon and Jamie Kiros and Jan Leike and Jane Park and Jason Kwon and Jason Phang and Jason Teplitz and Jason Wei and Jason Wolfe and Jay Chen and Jeff Harris and Jenia Varavva and Jessica Gan Lee and Jessica Shieh and Ji Lin and Jiahui Yu and Jiayi Weng and Jie Tang and Jieqi Yu and Joanne Jang and Joaquin Quinonero Candela and Joe Beutler and Joe Landers and Joel Parish and Johannes Heidecke and John Schulman and Jonathan Lachman and Jonathan McKay and Jonathan Uesato and Jonathan Ward and Jong Wook Kim and Joost Huizinga and Jordan Sitkin and Jos Kraaijeveld and Josh Gross and Josh Kaplan and Josh Snyder and Joshua Achiam and Joy Jiao and Joyce Lee and Juntang Zhuang and Justyn Harriman and Kai Fricke and Kai Hayashi and Karan Singhal and Katy Shi and Kavin Karthik and Kayla Wood and Kendra Rimbach and Kenny Hsu and Kenny Nguyen and Keren Gu-Lemberg and Kevin Button and Kevin Liu and Kiel Howe and Krithika Muthukumar and Kyle Luther and Lama Ahmad and Larry Kai and Lauren Itow and Lauren Workman and Leher Pathak and Leo Chen and Li Jing and Lia Guy and Liam Fedus and Liang Zhou and Lien Mamitsuka and Lilian Weng and Lindsay McCallum and Lindsey Held and Long Ouyang and Louis Feuvrier and Lu Zhang and Lukas Kondraciuk and Lukasz Kaiser and Luke Hewitt and Luke Metz and Lyric Doshi and Mada Aflak and Maddie Simens and Madelaine Boyd and Madeleine Thompson and Marat Dukhan and Mark Chen and Mark Gray and Mark Hudnall and Marvin Zhang and Marwan Aljubeh and Mateusz Litwin and Matthew Zeng and Max Johnson and Maya Shetty and Mayank Gupta and Meghan Shah and Mehmet Yatbaz and Meng Jia Yang and Mengchao Zhong and Mia Glaese and Mianna Chen and Michael Janner and Michael Lampe and Michael Petrov and Michael Wu and Michele Wang and Michelle Fradin and Michelle Pokrass and Miguel Castro and Miguel Oom Temudo de Castro and Mikhail Pavlov and Miles Brundage and Miles Wang and Minal Khan and Mira Murati and Mo Bavarian and Molly Lin and Murat Yesildal and Nacho Soto and Natalia Gimelshein and Natalie Cone and Natalie Staudacher and Natalie Summers and Natan LaFontaine and Neil Chowdhury and Nick Ryder and Nick Stathas and Nick Turley and Nik Tezak and Niko Felix and Nithanth Kudige and Nitish Keskar and Noah Deutsch and Noel Bundick and Nora Puckett and Ofir Nachum and Ola Okelola and Oleg Boiko and Oleg Murk and Oliver Jaffe and Olivia Watkins and Olivier Godement and Owen Campbell-Moore and Patrick Chao and Paul McMillan and Pavel Belov and Peng Su and Peter Bak and Peter Bakkum and Peter Deng and Peter Dolan and Peter Hoeschele and Peter Welinder and Phil Tillet and Philip Pronin and Philippe Tillet and Prafulla Dhariwal and Qiming Yuan and Rachel Dias and Rachel Lim and Rahul Arora and Rajan Troll and Randall Lin and Rapha Gontijo Lopes and Raul Puri and Reah Miyara and Reimar Leike and Renaud Gaubert and Reza Zamani and Ricky Wang and Rob Donnelly and Rob Honsby and Rocky Smith and Rohan Sahai and Rohit Ramchandani and Romain Huet and Rory Carmichael and Rowan Zellers and Roy Chen and Ruby Chen and Ruslan Nigmatullin and Ryan Cheu and Saachi Jain and Sam Altman and Sam Schoenholz and Sam Toizer and Samuel Miserendino and Sandhini Agarwal and Sara Culver and Scott Ethersmith and Scott Gray and Sean Grove and Sean Metzger and Shamez Hermani and Shantanu Jain and Shengjia Zhao and Sherwin Wu and Shino Jomoto and Shirong Wu and Shuaiqi and Xia and Sonia Phene and Spencer Papay and Srinivas Narayanan and Steve Coffey and Steve Lee and Stewart Hall and Suchir Balaji and Tal Broda and Tal Stramer and Tao Xu and Tarun Gogineni and Taya Christianson and Ted Sanders and Tejal Patwardhan and Thomas Cunninghman and Thomas Degry and Thomas Dimson and Thomas Raoux and Thomas Shadwell and Tianhao Zheng and Todd Underwood and Todor Markov and Toki Sherbakov and Tom Rubin and Tom Stasi and Tomer Kaftan and Tristan Heywood and Troy Peterson and Tyce Walters and Tyna Eloundou and Valerie Qi and Veit Moeller and Vinnie Monaco and Vishal Kuo and Vlad Fomenko and Wayne Chang and Weiyi Zheng and Wenda Zhou and Wesam Manassra and Will Sheu and Wojciech Zaremba and Yash Patil and Yilei Qian and Yongjik Kim and Youlong Cheng and Yu Zhang and Yuchen He and Yuchen Zhang and Yujia Jin and Yunxing Dai and Yury Malkov},
      year={2024},
      eprint={2410.21276},
      archivePrefix={arXiv},
      primaryClass={cs.CL},
      url={https://arxiv.org/abs/2410.21276}, 
}

@inproceedings{yang2018hotpotqa,
  title={{HotpotQA}: A Dataset for Diverse, Explainable Multi-hop Question Answering},
  author={Yang, Zhilin and Qi, Peng and Zhang, Saizheng and Bengio, Yoshua and Cohen, William W. and Salakhutdinov, Ruslan and Manning, Christopher D.},
  booktitle={Conference on Empirical Methods in Natural Language Processing ({EMNLP})},
  year={2018}
}

@incollection{williams2013gradient,
  title={Gradient-based learning algorithms for recurrent networks and their computational complexity},
  author={Williams, Ronald J and Zipser, David},
  booktitle={Backpropagation},
  pages={433--486},
  year={2013},
  publisher={Psychology Press}
}

@article{zhu2025superposition,
  title   = {Reasoning by Superposition: A Theoretical Perspective on Chain of Continuous Thought},
  author  = {Hanlin Zhu and Shibo Hao and Zhiting Hu and Jiantao Jiao and Stuart Russell and Yuandong Tian},
  journal = {arXiv preprint arXiv:2505.12514},
  year    = {2025},
  url     = {https://arxiv.org/abs/2505.12514},
}

@inproceedings{
loshchilov2018decoupled,
title={Decoupled Weight Decay Regularization},
author={Ilya Loshchilov and Frank Hutter},
booktitle={International Conference on Learning Representations},
year={2019},
url={https://openreview.net/forum?id=Bkg6RiCqY7},
}

@article{yang2024parallelizing,
  title={Parallelizing linear transformers with the delta rule over sequence length},
  author={Yang, Songlin and Wang, Bailin and Zhang, Yu and Shen, Yikang and Kim, Yoon},
  journal={Advances in neural information processing systems},
  volume={37},
  pages={115491--115522},
  year={2024}
}

@inproceedings{wu2025parallel,
  title={Parallel continuous chain-of-thought with jacobi iteration},
  author={Wu, Haoyi and Teng, Zhihao and Tu, Kewei},
  booktitle={Proceedings of the 2025 Conference on Empirical Methods in Natural Language Processing},
  pages={914--926},
  year={2025}
}
